\documentclass[letterpaper]{article}

\usepackage{amsfonts}  
\usepackage{amsmath}
\usepackage{amssymb}
\usepackage{amsthm}
\usepackage{appendix}
\usepackage{authblk}
\usepackage{bbold}
\usepackage{booktabs}  
\usepackage{color}
\usepackage{courier}
\usepackage{enumitem}
\usepackage{environ}
\usepackage{etoolbox}
\usepackage{float}
\usepackage{fullpage}
\usepackage{graphicx}
\usepackage{helvet}
\usepackage{mathtools}
\usepackage{mdwlist}  
\usepackage{microtype}  
\usepackage{multirow}
\usepackage{natbib}
\usepackage{nicefrac}  
\usepackage{parskip}
\usepackage{subfigure}
\usepackage[T1]{fontenc}  
\usepackage{tabularx}
\usepackage{times}
\usepackage{url}  
\usepackage[utf8]{inputenc}  
\usepackage{verbatim}
\usepackage{wrapfig}
\usepackage{xspace}

\usepackage{algorithm}
\usepackage{algorithmic}

\usepackage{hyperref}  

\setlist{nosep}

\hypersetup{colorlinks = false, pdfborder = {0 0 0}}

\newif\ifdraft
\draftfalse
\newcommand{\TODO}[1]{
  \ifdraft
    \textcolor{red}{[TODO: #1]}\xspace
  \fi
}
\newcommand{\NOTE}[1]{
  \ifdraft
    \textcolor{blue}{[NOTE: #1]}\xspace
  \fi
}

\newcommand{\google}{Google\xspace}

\renewcommand{\paragraph}[1]{\textbf{#1}\;\xspace}

\renewcommand{\eqref}[1]{Equation~\ref{eq:#1}}
\newcommand{\secref}[1]{Section~\ref{sec:#1}}
\newcommand{\appref}[1]{Appendix~\ref{app:#1}}
\newcommand{\figref}[1]{Figure~\ref{fig:#1}}
\newcommand{\tabref}[1]{Table~\ref{tab:#1}}
\newcommand{\algref}[1]{Algorithm~\ref{alg:#1}}
\newcommand{\thmref}[1]{Theorem~\ref{thm:#1}}
\newcommand{\lemref}[1]{Lemma~\ref{lem:#1}}

\newcommand{\defref}[1]{Definition~\ref{def:#1}}

\newcommand{\figrefs}[3][and]{Figures~\ref{fig:#2} #1~\ref{fig:#3}}

\newcommand{\algrefs}[3][and]{Algorithms~\ref{alg:#2} #1~\ref{alg:#3}}
\newcommand{\thmrefs}[3][and]{Theorems~\ref{thm:#2} #1~\ref{thm:#3}}

\newcommand{\algrefss}[4][and]{Algorithms~\ref{alg:#2}, \ref{alg:#3} #1~\ref{alg:#4}}

\newcommand{\algrefsss}[5][and]{Algorithms~\ref{alg:#2}, \ref{alg:#3}, \ref{alg:#4} #1~\ref{alg:#5}}

\newtheoremstyle{mytheoremstyle}
  {\parsep} 
  {} 
  {\itshape} 
  {} 
  {\bfseries} 
  {.} 
  {0.5em} 
  {} 
\theoremstyle{mytheoremstyle}

\makeatletter
\newtheorem*{rep@theorem}{\rep@title}
\newcommand{\newreptheorem}[2]{\newenvironment{rep#1}[1]{\def\rep@title{#2 \ref{##1}}\begin{rep@theorem}}{\end{rep@theorem}}}
\makeatother

\newtheorem{theorem}{Theorem}
\newtheorem{lemma}{Lemma}
\newtheorem{corollary}{Corollary}
\newtheorem{definition}{Definition}
\newreptheorem{theorem}{Theorem}
\newreptheorem{lemma}{Lemma}
\newreptheorem{corollary}{Corollary}

\newif\ifreptheorem
\reptheoremfalse
\newif\ifshowproofs
\showproofsfalse

\newcommand{\replabel}[1]{
  \ifreptheorem
    \tag{\ref{#1}}
  \else
    \label{#1}
  \fi
}
\NewEnviron{thm}[1]{
  \makeatletter
  \ifcsname defined:thm:#1\endcsname
    \reptheoremtrue
    \begin{reptheorem}{thm:#1}\BODY\end{reptheorem}
    \reptheoremfalse
  \else
    \begin{theorem}\label{thm:#1}\BODY\end{theorem}
    \csgdef{defined:thm:#1}{}
  \fi
  \makeatother
}
\NewEnviron{lem}[1]{
  \makeatletter
  \ifcsname defined:lem:#1\endcsname
    \reptheoremtrue
    \begin{replemma}{lem:#1}\BODY\end{replemma}
    \reptheoremfalse
  \else
    \begin{lemma}\label{lem:#1}\BODY\end{lemma}
    \csgdef{defined:lem:#1}{}
  \fi
  \makeatother
}
\NewEnviron{cor}[1]{
  \makeatletter
  \ifcsname defined:cor:#1\endcsname
    \reptheoremtrue
    \begin{repcorollary}{cor:#1}\BODY\end{repcorollary}
    \reptheoremfalse
  \else
    \begin{corollary}\label{cor:#1}\BODY\end{corollary}
    \csgdef{defined:cor:#1}{}
  \fi
  \makeatother
}
\NewEnviron{prf}[1]{
  \ifshowproofs
    \begin{proof}\BODY\end{proof}\label{app:prf:#1}
  \else
    \begin{proof}In \appref{prf:#1}.\end{proof}
  \fi
}

\newenvironment{titled-paragraph}[1]{\textbf{#1:}}{}

\newcommand{\tensorflow}{TensorFlow\xspace}

\newcommand{\iid}{\emph{i.i.d.}\xspace}
\newcommand{\wrt}{w.r.t.\xspace}
\newcommand{\ie}{i.e.\xspace}
\newcommand{\eg}{e.g.\xspace}
\newcommand{\egcite}[1]{\citep[\eg][]{#1}}

\newcommand{\probability}{\ensuremath{\mathrm{Pr}}}
\newcommand{\expectation}{\ensuremath{\mathbb{E}}}

\newcommand{\indicator}{\ensuremath{\mathbf{1}}}

\newcommand{\defeq}{\ensuremath{\vcentcolon=}}

\newcommand{\N}{\ensuremath{\mathbb{N}}}

\newcommand{\R}{\ensuremath{\mathbb{R}}}

\newcommand{\minimize}[1][]{\ensuremath{\underset{#1}{\mathrm{minimize}}\;}}
\newcommand{\suchthat}[1][]{\ensuremath{\underset{#1}{\mathrm{s.t.}}\;}}

\newcommand{\inner}[2]{\ensuremath{\left\langle {#1}, {#2} \right\rangle}\xspace}

\newcommand{\norm}[1]{\ensuremath{\left\lVert {#1} \right\rVert}\xspace}

\newcommand{\abs}[1]{\ensuremath{\left\lvert {#1} \right\rvert}\xspace}

\DeclareMathOperator*{\argmin}{argmin}

\newcommand{\indices}[1]{\ensuremath{\left[#1\right]}}

\DeclareMathOperator{\range}{range}

\newcommand{\codecomment}[1]{\`//~\textit{#1}}
\newcounter{lineno}
\newenvironment{pseudocode}{\setcounter{lineno}{0}\begin{tabbing}\textbf{mm}\=mm\=mm\=mm\=mm\=\kill}{\end{tabbing}}
\newcommand{\codename}{\>}

\newcommand{\codeline}{\>\stepcounter{lineno}\textbf{\arabic{lineno}}\'\>}

\DeclareMathOperator{\fix}{fix}

\newcommand{\elementwiseproduct}{\odot}

\newcommand{\elementwiseexp}{\operatorname{.exp}}

\newcommand{\stochasticgrad}{\ensuremath{\Delta}\xspace}
\newcommand{\stochasticsubgrad}{\ensuremath{\check{\stochasticgrad}}\xspace}
\newcommand{\stochasticsupgrad}{\ensuremath{\hat{\stochasticgrad}}\xspace}

\newcommand{\parameters}{\ensuremath{\theta}\xspace}
\newcommand{\Parameters}{\ensuremath{\Theta}\xspace}
\newcommand{\multipliers}{\ensuremath{\lambda}\xspace}
\newcommand{\Multipliers}{\ensuremath{\Lambda}\xspace}
\newcommand{\matrixmultipliers}{\ensuremath{M}\xspace}
\newcommand{\Matrixmultipliers}{\ensuremath{\mathcal{M}}\xspace}

\newcommand{\Radius}{\ensuremath{R}\xspace}
\newcommand{\margin}{\ensuremath{\gamma}\xspace}
\newcommand{\approximation}{\ensuremath{\rho}\xspace}
\newcommand{\oracle}{\ensuremath{\mathcal{O}_{\approximation}}\xspace}
\newcommand{\coveringradius}{r}
\newcommand{\covering}{\ensuremath{C_{\coveringradius}}\xspace}
\newcommand{\strongconvexity}{\ensuremath{\mu}\xspace}
\newcommand{\lipschitz}{\ensuremath{L}\xspace}

\newcommand{\dataset}{\ensuremath{S}\xspace}
\newcommand{\traindataset}{\ensuremath{\dataset^{(\mathrm{train})}}\xspace}
\newcommand{\valdataset}{\ensuremath{\dataset^{(\mathrm{val})}}\xspace}
\newcommand{\datadistribution}{\ensuremath{\mathcal{D}}\xspace}

\newcommand{\outputParameters}{\hat{\Parameters}\xspace}
\newcommand{\coveringParameters}{\Parameters_{\covering}\xspace}

\newcommand{\traingeneralization}[1]{\ensuremath{\tilde{G}^{(\mathrm{train})}(#1)}\xspace}
\newcommand{\valgeneralization}[1]{\ensuremath{G^{(\mathrm{val})}(#1)}\xspace}

\newcommand{\numconstraints}{\ensuremath{m}\xspace}

\newcommand{\loss}{\ensuremath{\ell}}
\newcommand{\constraintloss}[1]{\ensuremath{\loss_{#1}}\xspace}
\newcommand{\objectiveloss}{\constraintloss{0}\xspace}
\newcommand{\proxyconstraintloss}[1]{\ensuremath{\tilde{\loss}_{#1}}\xspace}
\newcommand{\lagrangian}{\ensuremath{\mathcal{L}}\xspace}
\newcommand{\empiricallagrangian}{\ensuremath{\hat{\mathcal{L}}}\xspace}

\newcommand{\bound}[1]{\ensuremath{B_{#1}}\xspace}

\title{Training Well-Generalizing Classifiers for Fairness Metrics and Other Data-Dependent Constraints}
\author[1]{Andrew Cotter}
\author[1]{Maya Gupta}
\author[1]{Heinrich Jiang}
\author[2]{Nathan Srebro}
\author[3]{Karthik Sridharan}
\author[1]{Serena Wang}
\author[2]{Blake Woodworth}
\author[4]{Seungil You}
\affil[1]{Google AI}
\affil[2]{Toyota Technological Institute at Chicago}
\affil[3]{Cornell University}
\affil[4]{Kakao Mobility}

\begin{document}

\maketitle

\begin{abstract}
Classifiers can be trained with data-dependent constraints to satisfy fairness
goals, reduce churn, achieve a targeted false positive rate, or other policy
goals.
We study the generalization performance for such constrained optimization
problems, in terms of how well the constraints are satisfied at evaluation
time, given that they are satisfied at training time. To improve generalization
performance, we frame the problem as a two-player game where one player
optimizes the model parameters on a training dataset, and the other player
enforces the constraints on an independent validation dataset.
We build on recent work in two-player constrained optimization to show that if
one uses this two-dataset approach, then constraint generalization can be
significantly improved.
As we illustrate experimentally, this approach works not only in theory, but
also in practice.

\end{abstract}

\section{Introduction}\label{sec:introduction}

It is useful to train classifiers with data-dependent constraints in order to
achieve certain guarantees on the training set, such as statistical parity or
other fairness guarantees, specified recall, or a desired positive
classification
rate~\egcite{Scott:2005,Zafar:2015,Goh:2016,Woodworth:2017,Narasimhan:2018}).
However, a key question is whether the achieved constraints will
\emph{generalize}. For example: will a classifier trained to produce $80\%$
statistical parity on training examples still achieve $80\%$ statistical parity
at evaluation time?

Unfortunately, the answer is ``not quite.'' Because such constraints are
data-dependent, overfitting can occur, and constraints that were satisfied on
the training set should be expected to be slightly violated on an \iid test
set. This is particularly problematic in the context of fairness constraints,
which will typically be chosen based on real-world requirements (\eg the $80\%$
rule of some US laws~\citep{Biddle:2005,Vuolo:2013,Zafar:2015,Hardt:2016}). In
this paper, we investigate how well constraints generalize, and propose
algorithms to improve the generalization of constraints to new examples.

Specifically, we consider problems that minimize a loss function subject to
data-dependent constraints, expressed in terms of \emph{expectations} over a
data distribution $\datadistribution$:
\begin{align}
  \label{eq:constrained-problem} \min_{\parameters \in \Parameters} \; &
  \expectation_{x \sim \datadistribution}\left[ \objectiveloss \left(x;
  \parameters\right) \right] \\
  \notag \suchthat[\forall i \in \indices{\numconstraints}] & \expectation_{x \sim
  \datadistribution}\left[ \constraintloss{i}\left(x; \parameters\right)
  \right] \le 0
\end{align}
where $x \in \mathcal{X}$ is a feature vector, $\mathcal{D}$ is the data
distribution over $\mathcal{X}$, $\Parameters$ is a space of model parameters
for the function class of interest, and
$\objectiveloss,\constraintloss{1},\dots,\constraintloss{\numconstraints}:
\mathcal{X} \times \Parameters \rightarrow \R$ are loss functions associated
with the objective and the $\numconstraints$
constraints~\footnote{\tabref{notation}, in the appendix, summarizes our
notation.}. We do \emph{not} require these loss functions to be convex.
\appref{examples} contains two examples of how \eqref{constrained-problem} can
be used to express certain data-dependent constraints (see \citet{Goh:2016} and
\citet{Narasimhan:2018} for more).

One typically trains a classifier on a finite training set drawn from
$\mathcal{D}$, but the true goal is to satisfy constraints \emph{in expectation
over $\mathcal{D}$}, as in \eqref{constrained-problem}.
%
%
To this end, we build on a long line of prior work that treats constrained
optimization as a two-player
game~\egcite{Christiano:2011,Arora:2012,Rakhlin:2013,Kearns:2017,Narasimhan:2018,Agarwal:2018}.
In this setting, the first player optimizes the model parameters $\parameters$,
and the second player enforces the constraints, \eg using the Lagrangian
formulation:
\begin{equation}
  \label{eq:lagrangian}
  \lagrangian\left(\parameters, \multipliers\right) \defeq \expectation_{x \sim
  \datadistribution}\left[ \objectiveloss\left(x;\parameters\right) +
  \sum_{i=1}^{\numconstraints} \multipliers_i
  \constraintloss{i}\left(x;\parameters\right) \right]
\end{equation}
In practice, one would approximate the Lagrangian with a finite \iid training
sample from $\datadistribution$, and the first player would minimize over the
model parameters $\parameters \in \Parameters$ while the second player
maximizes over the Lagrange multipliers $\multipliers \in
\R_+^{\numconstraints}$.

Our key idea is to treat constrained optimization similarly to hyperparameter
optimization: just as one typically chooses hyperparameters based on a
validation set, instead of the training set, to improve classifier
generalization, we would like to choose the Lagrange multipliers on a
validation set to improve \emph{constraint} generalization. In ``inner''
optimizations we would, given a fixed $\multipliers$, minimize the empirical
Lagrangian on the training set. Then, in an ``outer'' optimization, we would
choose a $\multipliers$ that results in the constraints being satisfied on the
validation set.
Such an approach, could it be made to work, would not eliminate the constraint
generalization problem completely---hyperparameter overfitting~\egcite{Ng:1997}
is a real problem---but would mitigate it, since constraint generalization
would no longer depend on size of the training sample and the complexity of
$\Parameters$ (which could be extremely large, \eg for a deep neural network),
but rather on the size of the validation sample and the effective complexity of
$\R_+^{\numconstraints} \ni \multipliers$, which, being
$\numconstraints$-dimensional, is presumably much simpler than $\Parameters$.

While the above approach is intuitive, challenges arise when attempting to
analyze it. The most serious is that since $\parameters$ is chosen based on the
training set, and $\multipliers$ on the validation set, the
$\parameters$-player is minimizing a \emph{different function} than the
$\multipliers$-player is maximizing, so the corresponding two-player game is
\emph{non-zero-sum} (the players have different cost functions). To handle
this, we must depart from the typical Lagrangian formulation, but the key idea
remains: improving generalization by using a separate validation set to enforce
the constraints.

Fortunately, the recent work of \citet{Cotter:2018} gives a strategy for
dealing with a non-zero-sum game in the context of constrained supervised
learning. We adapt their approach to this new setting to give bounds on
constraint generalization that are agnostic to model complexity.
After some preliminary definitions in \secref{background}, in
\secref{algorithms} we present two algorithms for which we can provide
theoretical bounds.

In \secref{experiments}, we perform a set of experiments demonstrating that our
two-dataset approach successfully improves constraint generalization even when
our theoretical results do not hold. In other words, providing independent
datasets to the $\parameters$- and $\multipliers$-players seems to work well
\emph{as a heuristic} for improving constraint generalization.

\section{Related Work}\label{sec:related}

While several recent papers have proved generalization bounds for constrained
problems~\egcite{Goh:2016,Agarwal:2018,Donini:2018}, the problem of
\emph{improving} constraint generalization is a fairly new one, having, so far
as we know, only been previously considered in the work of
\citet{Woodworth:2017}, who handled generalization subject to ``equalized
odds'' constraints in the setting of~\citet{Hardt:2016}. Specifically, their
approach is to first learn a predictor on $\traindataset$, and then to learn a
``correction'' on $\valdataset$ to more tightly satisfy the fairness
constraints. The second stage requires estimating only a constant number of
parameters, and the final predictor consequently enjoys a generalization
guarantee for the fairness constraints which is independent of the predictor's
complexity, with only a modest penalty to the loss. However, their approach
relies heavily upon the structure of equalized odds constraints: it requires
that any classifier can be modified to satisfy the fairness constraints
\emph{and} have low loss on a validation set by tuning only a small number of
parameters.

\citet{Woodworth:2017}'s overall approach can be summarized as ``train a
complicated model on a training set, and then a simple correction on a
validation set''. If, as they show to be the case for equalized odds
constraints, the ``simple correction'' is capable of satisfying the constraints
without significantly compromising on quality, then this technique results in a
well-performing model for which the validation constraint generalization
depends not on the complexity of the ``complicated model'', but rather of that
of the ``simple correction''. In this paper, we extend \citet{Woodworth:2017}'s
two-dataset idea to work on data-dependent constraints \emph{in general}.

Our primary baseline is \citet{Agarwal:2018}'s recently-proposed algorithm for
fair classification using the Lagrangian formulation. Their proposal, like our
\algref{discrete}, uses an oracle to optimize \wrt \parameters (they use the
terminology ``best response''), and, like all of our algorithms, results in a
stochastic classifier. However, our setting differs slightly from theirs---they
focus on fair classification, while we work in the slightly more general
inequality-constrained setting (\eqref{constrained-problem}). For this reason,
in \appref{baseline} we provide an analysis of the Lagrangian formulation for
inequality constrained optimization.

\section{Background \& Definitions}\label{sec:background}

Our algorithms are based on the non-zero-sum two-player game proposed by
\citet{Cotter:2018}, which they call the ``proxy-Lagrangian'' formulation.
The key novelty of their approach is the use of ``proxy'' constraint losses,
which are essentially surrogate losses that are used by only \emph{one} of the
two players (the $\parameters$-player).
It is because the two players use different losses that their proposed game is
non-zero-sum. The motivation behind their work is that a surrogate might be
necessary when the constraint functions are non-differentiable or discontinuous
(\eg for fairness metrics, which typically constrain proportions, \ie linear
combinations of indicators), but the overall goal is still to satisfy the
original (non-surrogate) constraints.
Our work differs in that we use a non-zero-sum game to provide different
\emph{datasets} to the two players, rather than different \emph{losses}.

Despite this difference, the use of proxy-constraints is perfectly compatible
with our proposal, so we permit the approximation of each of our constraint
losses $\ell_i$ with a (presumably differentiable) upper-bound
$\tilde{\ell}_i$. These are used \emph{only} by the $\parameters$-player; the
$\multipliers$-player uses the original constraint losses.
The use of proxy constraint losses is entirely optional: one is free to choose
$\proxyconstraintloss{i} \defeq \constraintloss{i}$ for all $i$.
\begin{definition}
  \label{def:empirical-proxy-lagrangians}
  Let \traindataset and \valdataset be two random datasets each drawn \iid from
  a data distribution \datadistribution.
  Given proxy constraint losses $\proxyconstraintloss{i}\left(x;
  \parameters\right) \ge \constraintloss{i}\left(x; \parameters\right)$ for all
  $x \in \mathcal{X}$, $\parameters \in \Parameters$ and $i \in
  \indices{\numconstraints}$, the empirical proxy-Lagrangians
  $\empiricallagrangian_{\parameters},\empiricallagrangian_{\multipliers} :
  \Parameters \times \Multipliers \rightarrow \R$ of
  \eqref{constrained-problem} are:
  \begin{align*}
    \empiricallagrangian_{\parameters}\left(\parameters, \multipliers\right)
    \defeq& \frac{1}{\abs{\traindataset}} \sum_{x \in \traindataset} \left(
    \multipliers_1 \objectiveloss \left( x; \parameters\right) +
    \sum_{i=1}^{\numconstraints} \multipliers_{i+1}
    \proxyconstraintloss{i}\left(x; \parameters\right) \right) \\
    \empiricallagrangian_{\multipliers}\left(\parameters, \multipliers\right)
    \defeq& \frac{1}{\abs{\valdataset}} \sum_{x \in \valdataset}
    \sum_{i=1}^{\numconstraints} \multipliers_{i+1} \constraintloss{i}\left(x;
    \parameters\right)
  \end{align*}
  where $\Multipliers \defeq \Delta^{\numconstraints+1}$ is the
  $\left(\numconstraints+1\right)$-dimensional simplex.
\end{definition}
The difference between the above, and Definition 2 of \citet{Cotter:2018}, is
that $\empiricallagrangian_{\parameters}$ is an empirical average over the
training set, while $\empiricallagrangian_{\multipliers}$ is over the
validation set.
The $\parameters$-player seeks to minimize $\empiricallagrangian_{\parameters}$
over $\parameters$, while the $\multipliers$-player seeks to maximize
$\empiricallagrangian_{\multipliers}$ over $\multipliers$.
In words, the $\multipliers$-player will attempt to satisfy the \emph{original}
constraints on the \emph{validation} set by choosing how much to penalize the
\emph{proxy} constraints on the \emph{training} set.
%

\subsection{Generalization}\label{sec:background:generalization}

Our ultimate interest is in generalization, and our bounds will be expressed in
terms of both the training and validation generalization errors, defined as
follows:
\begin{definition}
  \label{def:generalization}
  Define the training generalization error $\traingeneralization{\Parameters}$
  such that:
  \begin{equation*}
    \abs{ \expectation_{x \sim \datadistribution}\left[ \loss\left(x,
    \parameters\right) \right] - \frac{1}{\abs{\traindataset}} \sum_{x \in
    \traindataset} \loss\left(x, \parameters\right) } \le
    \traingeneralization{\Parameters}
  \end{equation*}
  for all $\parameters \in \Parameters$ and all $\loss \in
  \left\{\objectiveloss, \proxyconstraintloss{1}, \dots,
  \proxyconstraintloss{\numconstraints}\right\}$ (the objective and proxy
  constraint losses, but not the original constraint losses).

  Likewise, define the validation generalization error
  $\valgeneralization{\outputParameters}$ to satisfy the analogous inequality
  in terms of $\valdataset$:
  \begin{equation*}
    \abs{ \expectation_{x \sim \datadistribution}\left[ \loss\left(x,
    \parameters\right) \right] - \frac{1}{\abs{\valdataset}} \sum_{x \in
    \valdataset} \loss\left(x, \parameters\right) } \le
    \valgeneralization{\outputParameters}
  \end{equation*}
  for all $\parameters \in \outputParameters \subseteq \Parameters$ and all
  $\loss \in \left\{\constraintloss{1}, \dots,
  \constraintloss{\numconstraints}\right\}$ (the original constraint losses,
  but not the objective or proxy constraint losses).
\end{definition}
Throughout this paper, $\outputParameters \defeq
\{\parameters^{(1)},\dots,\parameters^{(T)}\} \subseteq \Parameters$ is the set
of $T$ iterates found by one of our proposed algorithms. Each of our guarantees
will be stated for a particular stochastic model $\bar{\parameters}$ supported
on $\outputParameters$ (\ie $\bar{\parameters}$ is a distribution over
$\outputParameters$), instead of for a single deterministic $\parameters \in
\outputParameters$. Notice that the above definitions of
$\traingeneralization{\Parameters}$ and $\valgeneralization{\outputParameters}$
also apply to such stochastic models: by the triangle inequality, if every
$\parameters \in \outputParameters$ generalizes well, then any
$\bar{\parameters}$ supported on $\outputParameters$ generalizes equally well,
in expectation.

\section{Algorithms}\label{sec:algorithms}

We seek a solution that (i) is nearly-optimal, (ii) nearly-feasible, and (iii)
generalizes well on the constraints. The optimality and feasibility goals were
already tackled by \citet{Cotter:2018} in the context of the proxy-Lagrangian
formulation of \defref{empirical-proxy-lagrangians}. They proposed having the
$\parameters$-player minimize ordinary external regret, and the
$\multipliers$-player minimize swap regret using an algorithm based on
\citet{Gordon:2008}. Rather than finding a \emph{single} solution (a pure
equilibrium of \defref{empirical-proxy-lagrangians}), they found a
\emph{distribution} over solutions (a mixed equilibrium). Our proposed approach
follows this same pattern, but we build on top of it to address challenge
(iii): generalization.

To this end, we draw inspiration from \citet{Woodworth:2017} (see
\secref{related}), and isolate the constraints from the complexity of
$\Parameters$ by using two independent datasets: $\traindataset$ and
$\valdataset$. The ``training'' dataset will be used to choose a good set of
model parameters $\parameters$, and the ``validation'' dataset to choose
$\multipliers$, and thereby impose the constraints. Like
\citet{Woodworth:2017}, the resulting constraint generalization bounds will be
\emph{independent} of the complexity of the function class.

We'll begin, in \secref{algorithms:discrete}, by proposing and analyzing an
oracle-based algorithm that improves generalization by discretizing the
candidate set, but makes few assumptions (not even convexity). Next, in
\secref{algorithms:continuous}, we give an algorithm that is more
``realistic''---there is no oracle, and no discretization---but requires
stronger assumptions, including strong convexity of the objective and
proxy-constraint losses $\objectiveloss, \proxyconstraintloss{1}, \dots,
\proxyconstraintloss{\numconstraints}$ (but \emph{not} of the original
constraint losses $\constraintloss{1}, \dots,
\constraintloss{\numconstraints}$).
%

In \secref{experiments}, we will present and perform experiments on simplified
``practical'' algorithms with no guarantees, but that incorporate our key idea:
having the $\multipliers$-player use an independent validation set.

\subsection{Covering-based Algorithm}\label{sec:algorithms:discrete}

\begin{algorithm*}[t]

\begin{pseudocode}
\codename $\mbox{DiscreteTwoDataset}\left( \empiricallagrangian_{\parameters}, \empiricallagrangian_{\multipliers} : \Parameters \times \Delta^{\numconstraints+1} \rightarrow \R, \oracle : \left(\Parameters \rightarrow \R\right) \rightarrow \Parameters, \covering \subseteq \R^{\numconstraints+1}, T \in \N, \eta_{\multipliers} \in \R_+ \right)$: \\
\codeline Initialize $\matrixmultipliers^{(1)} \in \R^{\left(\numconstraints + 1\right) \times \left(\numconstraints + 1\right)}$ with $\matrixmultipliers_{i,j} = 1 / \left(\numconstraints+1\right)$ \\
\codeline For $t \in \indices{T}$: \\
\codeline \> Let $\multipliers^{(t)} = \fix \matrixmultipliers^{(t)}$ \codecomment{Fixed point of $\matrixmultipliers^{(t)}$, \ie a stationary distribution} \\
\codeline \> Let $\tilde{\multipliers}^{(t)} = \argmin_{\tilde{\multipliers} \in \covering} \norm{\multipliers^{(t)} - \tilde{\multipliers}}_1$ \codecomment{Discretization to closest point in $\covering$} \\
\codeline \> Let $\parameters^{(t)} = \oracle\left( \empiricallagrangian_{\parameters}\left(\cdot,\tilde{\multipliers}^{(t)}\right) \right)$ \\
\codeline \> Let $\stochasticsupgrad^{(t)}_{\multipliers}$ be a supergradient of $\empiricallagrangian_{\multipliers}\left(\parameters^{(t)},\multipliers^{(t)}\right)$ \wrt $\multipliers$ \\
\codeline \> Update $\tilde{\matrixmultipliers}^{(t+1)} = \matrixmultipliers^{(t)} \elementwiseproduct \elementwiseexp\left( \eta_{\multipliers} \stochasticsupgrad^{(t)}_{\multipliers} \left( \multipliers^{(t)} \right)^T \right)$ \codecomment{$\elementwiseproduct$ and $\elementwiseexp$ are element-wise} \\
\codeline \> Project $\matrixmultipliers^{(t+1)}_{:,i} = \tilde{\matrixmultipliers}^{(t+1)}_{:,i} / \norm{\tilde{\matrixmultipliers}^{(t+1)}_{:,i}}_1$ for $i\in\indices{\numconstraints+1}$ \codecomment{Column-wise projection \wrt KL divergence} \\
\codeline Return $\parameters^{(1)},\dots,\parameters^{(T)}$ and $\multipliers^{(1)},\dots,\multipliers^{(T)}$
\end{pseudocode}

\caption{
  Finds an approximate equilibrium of the empirical proxy-Lagrangian game
  (\defref{empirical-proxy-lagrangians}), with \thmref{discrete} being its
  convergence and generalization guarantee. This is essentially a discretized
  version of Algorithm 4 of \citet{Cotter:2018}---like that algorithm,
  because of its dependence on an oracle, this algorithm does \emph{not} require
  convexity.
  Here, $\oracle$ is a deterministic Bayesian optimization oracle
  (\defref{oracle}), and $\covering$ is a radius-$\coveringradius$ (external)
  covering of $\Multipliers \defeq \Delta^{\numconstraints+1}$ \wrt the
  $1$-norm.
  The $\parameters$-player uses oracle calls to approximately minimize
  $\empiricallagrangian_{\parameters}\left(\cdot,\multipliers\right)$, while
  the $\multipliers$-player uses a swap-regret minimizing algorithm in the
  style of \citet{Gordon:2008}, using the left-stochastic state matrices
  $\matrixmultipliers^{(t)} \in
  \R^{(\numconstraints+1)\times(\numconstraints+1)}$.
}

\label{alg:discrete}

\end{algorithm*}

The simplest way to attack the generalization problem, and the first that we
propose, is to \emph{discretize} the space of allowed $\multipliers$s, and
associate each $\multipliers$ with a unique $\parameters \in \Parameters$,
where this association is based \emph{only} on the training set.
If the set of discretized $\multipliers$s is sufficiently small, then the set
of discretized $\parameters$s will likewise be small, and since it was chosen
independently of the validation set, its validation performance will generalize
well.

Specifically, we take $\covering$ to be a radius-$\coveringradius$ (external)
covering of $\Multipliers \defeq \Delta^{\numconstraints+1}$ \wrt the $1$-norm.
The set of allowed $\multipliers$s is exactly the covering centers, while,
following \citet{Chen:2017}, \citet{Agarwal:2018} and \citet{Cotter:2018}, the
associated $\parameters$s are found using an approximate Bayesian optimization
oracle:
%
%
\begin{definition}
  \label{def:oracle}
  A $\approximation$-approximate Bayesian optimization oracle is a function
  $\oracle : \left(\Parameters \rightarrow \R\right) \rightarrow \Parameters$
  for which:
  \begin{equation*}
    f\left( \oracle\left(f\right) \right) \le \inf_{\parameters^* \in
    \Parameters} f\left(\parameters^*\right) + \approximation
  \end{equation*}
  for any $f : \Parameters \rightarrow \R$ that can be written as
    $f\left(\parameters\right) = \empiricallagrangian_{\parameters}\left(
    \parameters, \multipliers \right)$ for some $\multipliers$. Furthermore,
    every time it is given the same $f$, $\oracle$ will return the same
    $\parameters$ (\ie it is deterministic).
\end{definition}
We will take the discretized set of $\parameters$s to be the oracle solutions
corresponding to the covering centers, \ie $\coveringParameters \defeq \{
\oracle( \empiricallagrangian_{\parameters}( \cdot, \tilde{\multipliers} )) :
\tilde{\multipliers} \in \covering \}$. The proof of the upcoming theorem shows
that if the radius parameter $\coveringradius$ is sufficiently small, then for
any achievable objective function value and corresponding constraint
violations, there will be a $\parameters \in \coveringParameters$ that is
almost as good. Hence, despite the use of discretization, we will still be able
to find a nearly-optimal and nearly-feasible solution.  Additionally, since the
set of discretized classifiers is finite, we can apply the standard
generalization bound for a finite function class, which will be tightest when
we take $\coveringradius$ to be as large as possible while still satisfying our
optimality and feasibility requirements.

\algref{discrete} combines our proposed discretization with the oracle-based
proxy-Lagrangian optimization procedure proposed by \citet{Cotter:2018}. As
desired, it finds a sequence of solutions $\outputParameters \defeq
\{\parameters^{(1)},\dots,\parameters^{(T)}\}$ for which it is possible to
bound $\valgeneralization{\outputParameters}$ \emph{independently} of the
complexity of the function class parameterized by $\Parameters$, \emph{and}
finds a random parameter vector $\bar{\parameters}$ supported on
$\outputParameters$ that is nearly-optimal and nearly-feasible.
\begin{thm}{discrete}
  Given any $\epsilon > 0$, there exists a covering $\covering$ such that, if
  we take $T \ge 4 \bound{\stochasticgrad}^2 \left(\numconstraints+1\right)
  \ln\left(\numconstraints+1\right) / \epsilon^2$ and $\eta_{\multipliers} =
  \sqrt{\left(\numconstraints+1\right) \ln\left(\numconstraints+1\right) / T
  \bound{\stochasticgrad}^2}$,
  where $\bound{\stochasticgrad} \ge \max_{t \in \indices{T}}\norm{
  \stochasticgrad_{\multipliers}^{(t)} }_{\infty}$ is a bound on the gradients,
  then the following hold, where $\outputParameters \defeq \left\{
  \parameters^{(1)}, \dots, \parameters^{(T)} \right\}$ is the set of results
  of \algref{discrete}.

  \begin{titled-paragraph}{Optimality and Feasibility}
  Let $\bar{\parameters}$ be a random variable taking values from
  $\outputParameters$, defined such that $\bar{\parameters} =
  \parameters^{(t)}$ with probability $\multipliers^{(t)}_1 / \sum_{s=1}^T
  \multipliers^{(s)}_1$, and let $\bar{\multipliers} \defeq \left(\sum_{t=1}^T
  \multipliers^{(t)}\right) / T$.
  Then $\bar{\parameters}$ is nearly-optimal in expectation:
  \begin{align}
    \replabel{eq:discrete:optimality}
    \expectation_{\bar{\parameters}, x \sim \datadistribution}\left[
    \objectiveloss\left(x; \bar{\parameters}\right) \right] \le &
    \inf_{\parameters^* \in \Parameters : \forall i . \expectation_{x \sim
    \datadistribution}\left[ \proxyconstraintloss{i}\left(x;
    \parameters^*\right) \right] \le 0} \expectation_{x \sim
    \datadistribution}\left[ \objectiveloss\left(x; \parameters^* \right)
    \right] \\
    \notag & + \frac{1}{\bar{\multipliers}_1} \left( \approximation + 2
    \epsilon + 2 \traingeneralization{\Parameters} +
    \valgeneralization{\outputParameters} \right)
  \end{align}
  and nearly-feasible:
  \begin{equation}
    \replabel{eq:discrete:feasibility}
    \max_{i \in \indices{\numconstraints}} \expectation_{\bar{\parameters}, x
    \sim \datadistribution}\left[ \constraintloss{i}\left(x;
    \bar{\parameters}\right) \right] \le \frac{\epsilon}{\bar{\multipliers}_1}
    + \valgeneralization{\outputParameters}
  \end{equation}
  Additionally, if there exists a $\parameters' \in \Parameters$ that satisfies
  all of the constraints with margin $\margin$ (\ie $\expectation_{x \sim
  \datadistribution}\left[ \constraintloss{i}\left(x; \parameters'\right)
  \right] \le -\margin$ for all $i \in \indices{\numconstraints}$), then:
  \begin{equation}
    \replabel{eq:discrete:lambda-bound}
    \bar{\multipliers}_1 \ge \frac{1}{\margin + \bound{\objectiveloss}} \left(
    \margin - \approximation - 2 \epsilon - 2 \traingeneralization{\Parameters}
    - \valgeneralization{\outputParameters} \right)
  \end{equation}
  where $\bound{\objectiveloss} \ge \sup_{\parameters \in \Parameters}
  \expectation_{x \sim \datadistribution}\left[ \objectiveloss\left(x;
  \parameters\right) \right] - \inf_{\parameters \in \Parameters}
  \expectation_{x \sim \datadistribution}\left[ \objectiveloss\left(x;
  \parameters\right) \right]$ is a bound on the range of the objective loss.
  \end{titled-paragraph}

  \begin{titled-paragraph}{Generalization}
  With probability $1-\delta$ over the sampling of $\valdataset$:
  \begin{equation}
    \replabel{eq:discrete:generalization-bound}
    \valgeneralization{\outputParameters} < \bound{\loss} \sqrt{ \frac{
      \numconstraints \ln \left( 10 \bound{\tilde{\loss}} / \epsilon \right) +
      \ln \left( 2 \numconstraints / \delta \right) }{2 \abs{\valdataset}} }
  \end{equation}
  where $\bound{\tilde{\loss}} \ge \abs{ \ell\left(x, \parameters\right) }$ for
  all $\ell \in \left\{ \objectiveloss, \proxyconstraintloss{1}, \dots,
  \proxyconstraintloss{\numconstraints} \right\}$, and
  $\bound{\loss} \ge \max_{i \in \indices{\numconstraints}} \left( b_i - a_i
  \right)$ assuming that the range of each $\constraintloss{i}$ is the interval
  $\left[ a_i, b_i \right]$.
  \end{titled-paragraph}
\end{thm}
\begin{prf}{discrete}
  The particular values we choose for $T$ and $\eta_{\multipliers}$ come from
  \lemref{discrete-convergence}, taking $\coveringradius = \epsilon / 2
  \bound{\tilde{\loss}}$, $\epsilon_{\parameters} = \approximation + 2
  \coveringradius \bound{\tilde{\loss}} = \approximation + \epsilon$, and
  $\epsilon_{\multipliers} = \epsilon$.
  The optimality and feasibility results then follow from
  \thmref{dataset-suboptimality}.

  For the bound on $\valgeneralization{\outputParameters}$, notice that by
  \lemref{covering-number}, there exists a radius-$\coveringradius$ covering
  $\covering$ \wrt the $1$-norm with $\abs{\covering} \le \left(5 /
  \coveringradius\right)^{\numconstraints} = \left(10 \bound{\tilde{\loss}} /
  \epsilon\right)^{\numconstraints}$. Substituting this, and the definition of
  $\coveringradius$, into the bound of \lemref{discrete-generalization} yields
  the claimed bound.
\end{prf}

\begin{table*}[t]

\centering

\caption{
  Simplified comparison of our suboptimality and infeasibility bounds
  (\thmrefs{discrete}{continuous}) to those for the Lagrangian formulation
  trained only on $\traindataset$.
  The ``one-dataset'' row is the result of an analysis of this one-dataset
  Lagrangian approach (essentially the same algorithm as
  \citet{Agarwal:2018}---see \appref{baseline} for details).
  The ``two-dataset'' row contains the results for our algorithms, which use
  the proxy-Lagrangian formulation on two independent datasets.
  In both cases, $\varepsilon$ measures how far the sequence of iterates is
  from being the appropriate type of equilibrium ($\varepsilon = \approximation
  + 2 \epsilon$ for \thmref{discrete}, and $\varepsilon = 2 \epsilon$ for
  \thmref{continuous}).
  The big-Os absorb only constants which are properties of the constrained
  problem (\eqref{constrained-problem}) and choice of proxy-constraint losses:
  $\margin$ and $\bound{\objectiveloss}$.
  The key difference is in the ``Infeasibility'' column: our proposal depends
  on $\valgeneralization{\outputParameters}$---which we bound independently of
  the model complexity (see \tabref{generalization})---rather than
  $\traingeneralization{\Parameters}$.
}

\label{tab:bounds}

\begin{tabular}{r|cc|c}
  \toprule
  \textbf{\# Datasets} & \textbf{Suboptimality} & \textbf{Infeasibility} & \textbf{Assuming} \\
  \midrule
  \textbf{One} &
  $O\left( \varepsilon + \traingeneralization{\Parameters} \right)$ &
  $O\left( \varepsilon + \traingeneralization{\Parameters} \right)$ &
  $\traingeneralization{\Parameters} \le \margin / 2$~\footnotemark \\
  %
  \textbf{Two} &
  $O\left( \varepsilon + \traingeneralization{\Parameters} +
  \valgeneralization{\outputParameters} \right)$ &
  $O\left( \varepsilon + \valgeneralization{\outputParameters} \right)$ &
  $\varepsilon + 2 \traingeneralization{\Parameters} +
  \valgeneralization{\outputParameters} \le \margin / 2$ \\
  \bottomrule
\end{tabular}

\end{table*}
\footnotetext{This condition could be removed by defining the feasibility
margin $\margin$ in terms of $\traindataset$ instead of $\datadistribution$,
causing $\margin$ to depend on the particular training sample, instead of being
solely a property of the constrained problem and choice of proxy-constraint
losses.}

\begin{table*}[t]

\centering

\caption{
  Comparison of the standard Rademacher complexity-based generalization bound
  (of the function class $\mathcal{F}$ parameterized by $\Parameters$) to our
  bounds on $\valgeneralization{\outputParameters}$. All bounds hold with
  probability $1 - \delta$, and we assume that $\abs{\traindataset} \propto n$
  and $\abs{\valdataset} \propto n$ (\eg if the data is split 50/50).
  The big-Os absorb only constants which are properties of the constrained
  problem (\eqref{constrained-problem}) and choice of proxy-constraint losses:
  $\bound{\loss}$, $\bound{\tilde{\loss}}$, $\lipschitz$ and
  $\strongconvexity$.
  For our algorithms, the validation generalization performance of the
  constraints is \emph{independent} of the model complexity.
}

\label{tab:generalization}

\begin{tabular}{ccc}
  \toprule
  \textbf{$\traingeneralization{\Parameters}$} & \textbf{$\valgeneralization{\outputParameters}$ (\thmref{discrete})} & \textbf{$\valgeneralization{\outputParameters}$ (\thmref{continuous})} \\
  \midrule
  $O\left( R_n\left(\mathcal{F}\right) + \sqrt{\frac{\ln(1/\delta)}{n}} \right)$ &
  $O\left( \sqrt{\frac{ \numconstraints \ln(1/\epsilon) + \ln(\numconstraints / \delta) }{n}} \right)$ &
  $O\left( \sqrt{\frac{ \numconstraints \ln n + \ln(\numconstraints / \delta) }{n}} + \epsilon \right)$ \\
  \bottomrule
\end{tabular}

\end{table*}

When reading the above result, it's natural to wonder about the role played by
$\bar{\multipliers}_1$. Recall that, unlike the Lagrangian formulation, the
proxy-Lagrangian formulation (\defref{empirical-proxy-lagrangians}) has a
weight $\multipliers_1$ associated with the \emph{objective}, in addition to
the $\numconstraints$ weights
$\multipliers_2,\dots,\multipliers_{\numconstraints+1}$ associated with the
constraints.
When the $i$th constraint is violated, the corresponding $\multipliers_{i+1}$
will grow, pushing $\multipliers_1$ towards zero. Conversely, when the
constraints are satisfied, $\multipliers_1$ will be pushed towards one.
In other words, the magnitude of $\multipliers_1$ encodes the
$\multipliers$-player's ``belief'' about the feasibility of the solution.
Just as, when using the Lagrangian formulation, Lagrange multipliers will tend
to be small on a feasible problem, the proxy-Lagrangian objective weight
$\bar{\multipliers}_1$ will tend to be large on a feasible problem, as shown by
\eqref{discrete:lambda-bound}, which guarantees that $\bar{\multipliers}_1$
will be bounded away from zero provided that there exists a margin-feasible
solution with a sufficiently large margin $\margin$.
In practice, of course, one need not rely on this lower bound: one can instead
simply inspect the behavior of the sequence of $\multipliers^{(t)}_1$'s during
optimization.

\eqref{discrete:lambda-bound} causes our results to be gated by the feasibility
margin. Specifically, it requires the training and validation datasets to
generalize well enough for $\approximation + 2 \epsilon + 2
\traingeneralization{\Parameters} + \valgeneralization{\outputParameters}$ to
stay within the feasibility margin $\margin$.
Past this critical threshold, $\bar{\multipliers}_1$ can be lower-bounded by a
constant, and can therefore be essentially ignored.
To get an intuitive grasp of this condition, notice that it is similar to
requiring $\margin$-margin-feasible solutions on the training dataset to
generalize well enough to also be margin-feasible (with a smaller margin) on
the validation dataset, and vice-versa.

\tabref{bounds} contains a comparison of our bounds, obtained with the
proxy-Lagrangian formulation and two datasets, versus bounds for the standard
Lagrangian on one dataset. The ``Assuming'' column contains a condition
resulting from the above discussion.
There are two key ways in which our results improve on those for the
one-dataset Lagrangian: (i) in the ``Infeasibility'' column, our approach
depends on $\valgeneralization{\outputParameters}$ instead of
$\traingeneralization{\Parameters}$, and (ii): as shown in
\tabref{generalization}, for our algorithms the generalization performance
$\valgeneralization{\outputParameters}$ of the constraints is bounded
\emph{independently} of the complexity of $\Parameters$.

It's worth emphasizing that this generalization bound (\tabref{generalization})
is distinct from the feasibility bound (the ``Infeasibility'' column of
\tabref{bounds}). When using our algorithms, testing constraint violations will
\emph{always} be close to the validation violations, regardless of the value of
$\bar{\multipliers}_1$. The ``Assuming'' column is only needed when asking
whether the validation violations are close to zero.

\subsection{Gradient-based Algorithm}\label{sec:algorithms:continuous}

\begin{algorithm*}[t]

\begin{pseudocode}
\codename $\mbox{ContinuousTwoDataset}\left( \empiricallagrangian_{\parameters}, \empiricallagrangian_{\multipliers} : \Parameters \times \Delta^{\numconstraints+1} \rightarrow \R, T_{\parameters}, T_{\multipliers} \in \N, \strongconvexity, \eta_{\multipliers} \in \R_+ \right)$: \\
\codeline Initialize $\matrixmultipliers^{(1)} \in \R^{\left(\numconstraints + 1\right) \times \left(\numconstraints + 1\right)}$ with $\matrixmultipliers_{i,j} = 1 / \left(\numconstraints+1\right)$ \\
\codeline For $t \in \indices{T_{\multipliers}}$: \\
\codeline \> Let $\multipliers^{(t)} = \fix \matrixmultipliers^{(t)}$ \codecomment{fixed point of $\matrixmultipliers^{(t)}$, \ie a stationary distribution} \\
\codeline \> For $s \in \indices{T_{\parameters}}$: \\
\codeline \> \> Initialize $\tilde{\parameters}^{(t,1)} = 0$ \codecomment{Assumes $0 \in \Parameters$} \\
\codeline \> \> Let $\stochasticsubgrad^{(t,s)}_{\parameters}$ be a subgradient of $\empiricallagrangian_{\parameters}\left(\tilde{\parameters}^{(t,s)},\multipliers^{(t)}\right)$ \wrt $\parameters$ \\
\codeline \> \> Update $\tilde{\parameters}^{(t,s+1)} = \Pi_{\Parameters}\left( \tilde{\parameters}^{(t,s)} - \stochasticsubgrad^{(t,s)}_{\parameters} / \strongconvexity s \right)$ \\
\codeline \> Define $\parameters^{(t)} \defeq \left( \sum_{s=1}^{T_{\parameters}} \tilde{\parameters}^{(t,s)} \right) / T_{\parameters}$ \\
\codeline \> Let $\stochasticgrad^{(t)}_{\multipliers}$ be a gradient of $\empiricallagrangian_{\multipliers}\left(\parameters^{(t)},\multipliers^{(t)}\right)$ \wrt $\multipliers$ \\
\codeline \> Update $\tilde{\matrixmultipliers}^{(t+1)} = \matrixmultipliers^{(t)} \elementwiseproduct \elementwiseexp\left( \eta_{\multipliers} \stochasticgrad^{(t)}_{\multipliers} \left( \multipliers^{(t)} \right)^T \right)$ \codecomment{$\elementwiseproduct$ and $\elementwiseexp$ are element-wise} \\
\codeline \> Project $\matrixmultipliers^{(t+1)}_{:,i} = \tilde{\matrixmultipliers}^{(t+1)}_{:,i} / \norm{\tilde{\matrixmultipliers}^{(t+1)}_{:,i}}_1$ for $i\in\indices{\numconstraints+1}$ \codecomment{Column-wise projection \wrt KL divergence} \\
\codeline Return $\parameters^{(1)},\dots,\parameters^{(T)}$ and $\multipliers^{(1)},\dots,\multipliers^{(T)}$
\end{pseudocode}

\caption{
  Finds an approximate equilibrium of the empirical proxy-Lagrangian game
  (\defref{empirical-proxy-lagrangians}) assuming that $\loss\left(x;
  \parameters\right)$ is $\strongconvexity$-strongly convex in $\parameters$
  for all $\loss \in \left\{ \objectiveloss, \proxyconstraintloss{1}, \dots,
  \proxyconstraintloss{\numconstraints} \right\}$ (the objective and proxy
  constraint losses, but \emph{not} the original constraint losses).
  \thmref{continuous} is its convergence and generalization guarantee.
  The $\parameters$-player uses gradient descent, while the
  $\multipliers$-player uses the same swap-regret minimizing procedure as
  \algref{discrete}.
}

\label{alg:continuous}

\end{algorithm*}

Aside from the unrealistic requirement for a Bayesian optimization oracle, the
main disadvantage of \algref{discrete} is that it relies on discretization. Our
next algorithm instead makes much stronger assumptions---strong convexity of
the objective and proxy constraint losses, and Lipschitz continuity of the
original constraint losses---enabling us to dispense with discretization
entirely in both the algorithm and the corresponding theorem statement.

The \emph{proof} of the upcoming theorem, however, still uses a covering. The
central idea is the same as before, with one extra step: thanks to strong
convexity, every (approximate) minimizer of
$\empiricallagrangian_{\parameters}(\cdot, \multipliers)$ is close to one of
the discretized parameter vectors $\parameters \in \coveringParameters$. Hence,
the set of such minimizers generalizes as well as $\coveringParameters$, plus
an additional term measuring the cost that we pay for approximating the
minimizers with elements of $\coveringParameters$.

The strong convexity assumption also enables us to replace the oracle call with
an explicit minimization procedure: gradient descent. The result is
\algref{continuous}, which, like \algref{discrete}, both finds a nearly-optimal
and nearly-feasible solution, \emph{and} enables us to bound
$\valgeneralization{\outputParameters}$ independently of the complexity of
$\Parameters$. Unlike \algref{discrete}, however, it is realistic enough to
permit a straightforward implementation.
\begin{thm}{continuous}
  Suppose that $\Parameters$ is compact and convex, and that $\loss\left(x;
  \parameters\right)$ is $\strongconvexity$-strongly convex in $\parameters$
  for all $\loss \in \left\{ \objectiveloss, \proxyconstraintloss{1}, \dots,
  \proxyconstraintloss{\numconstraints} \right\}$.
  Given any $\epsilon > 0$, if we take $T_{\parameters} \ge \left(
  \bound{\stochasticsubgrad}^2 / \strongconvexity \epsilon \right) \ln\left(
  \bound{\stochasticsubgrad}^2 / \strongconvexity \epsilon \right)$,
  $T_{\multipliers} \ge 4 \bound{\stochasticgrad}^2
  \left(\numconstraints+1\right) \ln\left(\numconstraints+1\right) /
  \epsilon^2$ and $\eta_{\multipliers} = \sqrt{\left(\numconstraints+1\right)
  \ln\left(\numconstraints+1\right) / T_{\multipliers}
  \bound{\stochasticgrad}^2}$,
  where $\bound{\stochasticgrad}$ is as in \thmref{discrete} and
  $\bound{\stochasticsubgrad} \ge \max_{s,t \in \indices{T_{\parameters}}
  \times \indices{T_{\multipliers}}}\norm{
  \stochasticsubgrad_{\parameters}^{(t,s)} }_{2}$ is a bound on the
  subgradients,
  then the following hold, where $\outputParameters \defeq \left\{
  \parameters^{(1)}, \dots, \parameters^{(T_{\multipliers})} \right\}$ is the
  set of results of \algref{discrete}.

  \begin{titled-paragraph}{Optimality and Feasibility}
  Let $\bar{\parameters}$ be a random variable taking values from
  $\outputParameters$, defined such that $\bar{\parameters} =
  \parameters^{(t)}$ with probability $\multipliers^{(t)}_1 / \sum_{s=1}^T
  \multipliers^{(s)}_1$, and let $\bar{\multipliers} \defeq \left(\sum_{t=1}^T
  \multipliers^{(t)}\right) / T_{\multipliers}$.
  Then $\bar{\parameters}$ is nearly-optimal in expectation:
  \begin{align*}
    %
    %
    \expectation_{\bar{\parameters}, x \sim \datadistribution}\left[
    \objectiveloss\left(x; \bar{\parameters}\right) \right] \le &
    \inf_{\parameters^* \in \Parameters : \forall i .  \expectation_{x \sim
    \datadistribution}\left[ \proxyconstraintloss{i}\left(x;
    \parameters^*\right) \right] \le 0} \expectation_{x \sim
    \datadistribution}\left[ \objectiveloss\left(x; \parameters^* \right)
    \right] \\
    %
    %
    & + \frac{1}{\bar{\multipliers}_1} \left( 2
    \epsilon + 2 \traingeneralization{\Parameters} +
    \valgeneralization{\outputParameters} \right)
  \end{align*}
  and nearly-feasible:
  \begin{equation*}
    %
    %
    \max_{i \in \indices{\numconstraints}} \expectation_{\bar{\parameters}, x
    \sim \datadistribution}\left[ \constraintloss{i}\left(x;
    \bar{\parameters}\right) \right] \le \frac{\epsilon}{\bar{\multipliers}_1}
    + \valgeneralization{\outputParameters}
  \end{equation*}
  Additionally, if there exists a $\parameters' \in \Parameters$ that satisfies
  all of the constraints with margin $\margin$ (\ie $\expectation_{x \sim
  \datadistribution}\left[ \constraintloss{i}\left(x; \parameters'\right)
  \right] \le -\margin$ for all $i \in \indices{\numconstraints}$), then:
  \begin{equation*}
    %
    %
    \bar{\multipliers}_1 \ge \frac{1}{\margin + \bound{\objectiveloss}} \left(
    \margin - 2 \epsilon - 2 \traingeneralization{\Parameters} -
    \valgeneralization{\outputParameters} \right)
  \end{equation*}
  where $\bound{\objectiveloss}$ is as in \thmref{discrete}.
  \end{titled-paragraph}

  \begin{titled-paragraph}{Generalization}
  If, in addition to the above requirements, $\loss\left(x; \parameters\right)$
  is $\lipschitz$-Lipschitz continuous in $\parameters$ for all $\loss \in
  \left\{ \constraintloss{1}, \dots, \constraintloss{\numconstraints}
  \right\}$, then with probability $1-\delta$ over the sampling of
  $\valdataset$:
  \begin{equation}
    \replabel{eq:continuous:generalization-bound}
    \valgeneralization{\outputParameters} <
    \bound{\loss} \sqrt{ \frac{ 2 \numconstraints }{\abs{\valdataset}}
    \max\left\{ 1, \ln\left( \frac{ 160 \lipschitz^2 \bound{\tilde{\loss}}
    \abs{\valdataset} }{ \numconstraints \strongconvexity \bound{\loss}^2 }
    \right) \right\} }
    %
    %
    + \bound{\loss} \sqrt{ \frac{ \ln \left( 2 \numconstraints / \delta
    \right) }{2 \abs{\valdataset}} }
    + 2 \lipschitz \epsilon \sqrt{ \frac{2}{\mu} }
  \end{equation}
  where $\bound{\tilde{\loss}}$ and $\bound{\loss}$ are as in
  \thmref{discrete}.
  \end{titled-paragraph}
\end{thm}
\begin{prf}{continuous}
  The particular values we choose for $T_{\parameters}$, $T_{\multipliers}$ and
  $\eta_{\multipliers}$ come from \lemref{continuous-convergence}, taking
  $\epsilon_{\parameters} = \epsilon_{\multipliers} = \epsilon$.
  The optimality and feasibility results then follow from
  \thmref{dataset-suboptimality}.

  For the bound on $\valgeneralization{\outputParameters}$, notice that
  by \lemref{covering-number}, there exists a radius-$\coveringradius$ external
  covering $\covering$ \wrt the $1$-norm with $\abs{\covering} \le
  \max\left\{1, \left(5 / \coveringradius\right)^{\numconstraints} \right\}$.
  Substituting into the bound of \lemref{continuous-generalization}:
  \begin{align*}
    \valgeneralization{\outputParameters} < &
    2 \lipschitz \sqrt{ \frac{4 \coveringradius
    \bound{\tilde{\loss}}}{\strongconvexity} } +
    \frac{2 \lipschitz \bound{\stochasticsubgrad}}{\strongconvexity} \sqrt{ \frac{1 + \ln
    T_{\parameters}}{T_{\parameters}} }
    + \bound{\loss} \sqrt{ \frac{ \numconstraints \max\left\{ 0, \ln\left( 5 /
    \coveringradius \right) \right\} + \ln \left( 2 \numconstraints / \delta
    \right) }{2 \abs{\valdataset}} } \\
    < &
    2 \lipschitz \sqrt{ \frac{4 \coveringradius
    \bound{\tilde{\loss}}}{\strongconvexity} }
    + \bound{\loss} \sqrt{ \frac{ \numconstraints \max\left\{ 0, \ln\left( 5 /
    \coveringradius \right) \right\} }{2 \abs{\valdataset}} }
    + \frac{2 \lipschitz \bound{\stochasticsubgrad}}{\strongconvexity} \sqrt{
      \frac{1 + \ln T_{\parameters}}{T_{\parameters}} }
    + \bound{\loss} \sqrt{ \frac{ \ln \left( 2 \numconstraints / \delta \right)
    }{2 \abs{\valdataset}} }
  \end{align*}
  Taking $\coveringradius \defeq \left( \numconstraints \strongconvexity
  \bound{\loss}^2 \right) / \left( 32 \lipschitz^2 \bound{\tilde{\loss}}
  \abs{\valdataset} \right)$:
  \begin{align*}
    \valgeneralization{\outputParameters} < &
    2 \bound{\loss} \sqrt{ \frac{ \numconstraints \max\left\{ 1, \ln\left( 5 /
    \coveringradius \right) \right\} }{2 \abs{\valdataset}} }
    + \frac{2 \lipschitz \bound{\stochasticsubgrad}}{\strongconvexity} \sqrt{
      \frac{1 + \ln T_{\parameters}}{T_{\parameters}} }
    + \bound{\loss} \sqrt{ \frac{ \ln \left( 2 \numconstraints / \delta \right)
    }{2 \abs{\valdataset}} } \\
    < &
    \bound{\loss} \sqrt{ \frac{ 2 \numconstraints }{\abs{\valdataset}}
    \max\left\{ 1, \ln\left( \frac{ 160 \lipschitz^2 \bound{\tilde{\loss}}
    \abs{\valdataset} }{ \numconstraints \strongconvexity \bound{\loss}^2 }
    \right) \right\} } \\
    & + \bound{\loss} \sqrt{ \frac{ \ln \left( 2 \numconstraints / \delta \right)
    }{2 \abs{\valdataset}} }
    + \frac{2 \lipschitz \bound{\stochasticsubgrad}}{\strongconvexity} \sqrt{
      \frac{1 + \ln T_{\parameters}}{T_{\parameters}} }
  \end{align*}
  Substituting the definition of $T_{\parameters}$ then yields the claimed result.
\end{prf}

The above optimality and feasibility guarantees are very similar to those of
\thmref{discrete}, as is shown in \tabref{bounds} (in which the only difference
is the definition of $\varepsilon$).
\algref{continuous}'s generalization bound
(\eqref{continuous:generalization-bound}) is more complicated than that of
\algref{discrete} (\eqref{discrete:generalization-bound}), but
\tabref{generalization} shows that the two are roughly comparable.
Hence, the overall theoretical performance of \algref{continuous} is very
similar to that of \algref{discrete}, and, while it does rely on stronger
assumptions, it neither uses discretization, nor does it require an oracle.

\section{Experiments}\label{sec:experiments}

\begin{algorithm*}[t]

\begin{pseudocode}
\codename $\mbox{PracticalTwoDataset}\left( \empiricallagrangian_{\parameters}, \empiricallagrangian_{\multipliers} : \Parameters \times \Delta^{\numconstraints+1} \rightarrow \R, T \in \N, \eta_{\parameters}, \eta_{\multipliers} \in \R_+ \right)$: \\
\codeline Initialize $\parameters^{(1)} = 0$ \codecomment{Assumes $0 \in \Parameters$} \\
\codeline Initialize $\matrixmultipliers^{(1)} \in \R^{\left(\numconstraints + 1\right) \times \left(\numconstraints + 1\right)}$ with $\matrixmultipliers_{i,j} = 1 / \left(\numconstraints+1\right)$ \\
\codeline For $t \in \indices{T}$: \\
\codeline \> Let $\multipliers^{(t)} = \fix \matrixmultipliers^{(t)}$ \codecomment{fixed point of $\matrixmultipliers^{(t)}$, \ie a stationary distribution} \\
\codeline \> Let $\stochasticsubgrad^{(t)}_{\parameters}$ be a stochastic subgradient of $\empiricallagrangian_{\parameters}\left(\parameters^{(t)},\multipliers^{(t)}\right)$ \wrt $\parameters$ \\
\codeline \> Let $\stochasticgrad^{(t)}_{\multipliers}$ be a stochastic gradient of $\empiricallagrangian_{\multipliers}\left(\parameters^{(t)},\multipliers^{(t)}\right)$ \wrt $\multipliers$ \\
\codeline \> Update $\parameters^{(t+1)} = \Pi_{\Parameters}\left( \parameters^{(t)} - \eta_{\parameters} \stochasticsubgrad^{(t)}_{\parameters} \right)$ \\
\codeline \> Update $\tilde{\matrixmultipliers}^{(t+1)} = \matrixmultipliers^{(t)} \elementwiseproduct \elementwiseexp\left( \eta_{\multipliers} \stochasticgrad^{(t)}_{\multipliers} \left( \multipliers^{(t)} \right)^T \right)$ \codecomment{$\elementwiseproduct$ and $\elementwiseexp$ are element-wise} \\
\codeline \> Project $\matrixmultipliers^{(t+1)}_{:,i} = \tilde{\matrixmultipliers}^{(t+1)}_{:,i} / \norm{\tilde{\matrixmultipliers}^{(t+1)}_{:,i}}_1$ for $i\in\indices{\numconstraints+1}$ \codecomment{Column-wise projection \wrt KL divergence} \\
\codeline Return $\parameters^{(1)},\dots,\parameters^{(T)}$ and $\multipliers^{(1)},\dots,\multipliers^{(T)}$
\end{pseudocode}

\caption{
  ``Practical'' algorithm for optimizing the empirical proxy-Lagrangian game
  (\defref{empirical-proxy-lagrangians}). This is essentially Algorithm 2 of
  \citet{Cotter:2018}, differing only in that it is applied to the two-dataset
  formulation of \defref{empirical-proxy-lagrangians}.
}

\label{alg:practical}

\end{algorithm*}

\begin{algorithm*}[t]

\begin{pseudocode}
\codename $\mbox{LagrangianTwoDataset}\left( \empiricallagrangian_{\parameters}, \empiricallagrangian_{\multipliers} : \Parameters \times \Delta^{\numconstraints} \rightarrow \R, T \in \N, \eta_{\parameters}, \eta_{\multipliers} \in \R_+ \right)$: \\
\codeline Initialize $\parameters^{(1)} = 0$, $\multipliers^{(1)} = 0$ \codecomment{Assumes $0 \in \Parameters$} \\
\codeline For $t \in \indices{T}$: \\
\codeline \> Let $\stochasticsubgrad^{(t)}_{\parameters}$ be a stochastic subgradient of $\lagrangian\left(\parameters^{(t)},\multipliers^{(t)}\right)$ \wrt $\parameters$ \\
\codeline \> Let $\stochasticgrad^{(t)}_{\multipliers}$ be a stochastic gradient of $\lagrangian\left(\parameters^{(t)},\multipliers^{(t)}\right)$ \wrt $\multipliers$ \\
\codeline \> Update $\parameters^{(t+1)} = \Pi_{\Parameters}\left( \parameters^{(t)} - \eta_{\parameters} \stochasticsubgrad^{(t)}_{\parameters} \right)$ \codecomment{Projected SGD updates \dots} \\
\codeline \> Update $\multipliers^{(t+1)} = \Pi_{\Multipliers}\left( \multipliers^{(t)} + \eta_{\multipliers} \stochasticgrad^{(t)}_{\multipliers} \right)$ \codecomment{\;\;\;\;\dots} \\
\codeline Return $\parameters^{(1)},\dots,\parameters^{(T)}$ and $\multipliers^{(1)},\dots,\multipliers^{(T)}$
\end{pseudocode}

\caption{
  ``Practical'' algorithm for optimizing a variant of the standard Lagrangian
  game, modified to support proxy constraints and two datasets. Here, instead
  of using the proxy-Lagrangian formulation of
  \defref{empirical-proxy-lagrangians}, we take
  $\empiricallagrangian_{\parameters} \defeq \expectation_{x \sim
  \traindataset}\left[ \objectiveloss(x;\parameters) +
  \sum_{i=1}^{\numconstraints} \multipliers_i \proxyconstraintloss{i}(x;
  \parameters) \right]$ and $\empiricallagrangian_{\multipliers} \defeq
  \expectation_{x \sim \valdataset}\left[ \objectiveloss(x;\parameters) +
  \sum_{i=1}^{\numconstraints} \multipliers_i \constraintloss{i}(x;
  \parameters) \right]$, with $\multipliers \in \Multipliers \defeq
  \R_+^{\numconstraints}$.
  Compared to \algref{practical}, this algorithm is further from those for
  which we can prove theoretical results (\algrefs{discrete}{continuous}), but
  is much closer to the Lagrangian-based approach of \citet{Agarwal:2018}. We
  include it to demonstrate that our two-dataset proposal works \emph{as a
  heuristic}.
}

\label{alg:lagrangian-practical}

\end{algorithm*}

While \secref{algorithms} has demonstrated the theoretical performance of
\algrefs{discrete}{continuous}, we believe that our proposed two-dataset
approach is useful \emph{as a heuristic} for improving constraint
generalization performance, even when one is not using a
theoretically-justified algorithm. For this reason, we experiment with two
``practical'' algorithms.
The first, \algref{practical}, is a bare-bones version of \algref{continuous},
in which $\parameters$ and $\multipliers$ are updated simultaneously using
stochastic updates, instead of in an inner and outer loop. This algorithm
implements our central idea---imposing constraints using an independent
validation dataset---without compromising on simplicity or speed.
The purpose of the second, \algref{lagrangian-practical}, is to explore how
well our two-dataset idea can be applied to the usual Lagrangian formulation.
For this algorithm, proxy-constraints and the use of two independent datasets
are essentially ``tacked on'' to the Lagrangian.
Neither of these algorithms enjoys the theoretical guarantees of
\secref{algorithms}, but, as we will see, both are still successful at
improving constraint generalization.

We present two sets of experiments, the first on simulated data, and the second
on real data. In both cases, each dataset is split into thee parts: training,
validation and testing. We compare our proposed two-dataset approach, in which
$\traindataset$ is the training dataset and $\valdataset$ is the validation
dataset, to the the natural baseline one-dataset approach of using the
\emph{union} of the training and validation sets to define \emph{both}
$\traindataset$ and $\valdataset$. Hence, both approaches ``see'' the same
total amount of data during training.

This difference between the data provided to the two algorithms
leads to a slight complication when reporting ``training'' error rates and
constraint violations. For the two-dataset approach, the former are reported on
$\traindataset$ (used to learn $\parameters$), and the latter on $\valdataset$
(used to learn $\multipliers$). For the baseline one-dataset algorithm, both
quantities are reported on the full dataset (\ie the union of the training and
validation sets). ``Testing'' numbers are always reported on the testing
dataset.

Our implementation uses \tensorflow, and is based on \citet{Cotter:2018}'s
open-source constrained optimization library. To avoid a hyperparameter search,
we replace the stochastic gradient updates of
\algrefs{practical}{lagrangian-practical} with ADAM~\citep{Kingma:2015}, using
the default parameters.
For both our two-dataset algorithm and the one-dataset baseline, the result of
training is a sequence of iterates $\parameters^{(1)},\dots,\parameters^{(T)}$,
but instead of keeping track of the full sequence, we only store a total of
$100$ evenly-spaced iterates for each run. Rather than using the weighted
predictor of \thmrefs{discrete}{continuous}, we use the ``shrinking'' procedure
of \citet{Cotter:2018} (see \appref{shrinking}) to find the \emph{best}
stochastic classifier supported on the sequence of $100$ iterates.

In all of our experiments, the objective and proxy constraint functions
$\objectiveloss,\proxyconstraintloss{1},\dots,\proxyconstraintloss{\numconstraints}$
are hinge upper bounds on the quantities of interest, while the \emph{original}
constraint functions
$\constraintloss{1},\dots,\constraintloss{\numconstraints}$ are precisely what
we claim to constrain (in these experiments, proportions, represented as linear
combinations of indicator functions).

\subsection{Simulated-data Experiments}\label{sec:experiments:simulated}

\begin{figure*}[t]

\centering

\begin{tabular}{cc}
  \multicolumn{2}{c}{\textbf{Linear Models}} \\
  \textbf{0/1 error} & \textbf{Constraint violation} \\
  \includegraphics[width=0.40\textwidth]{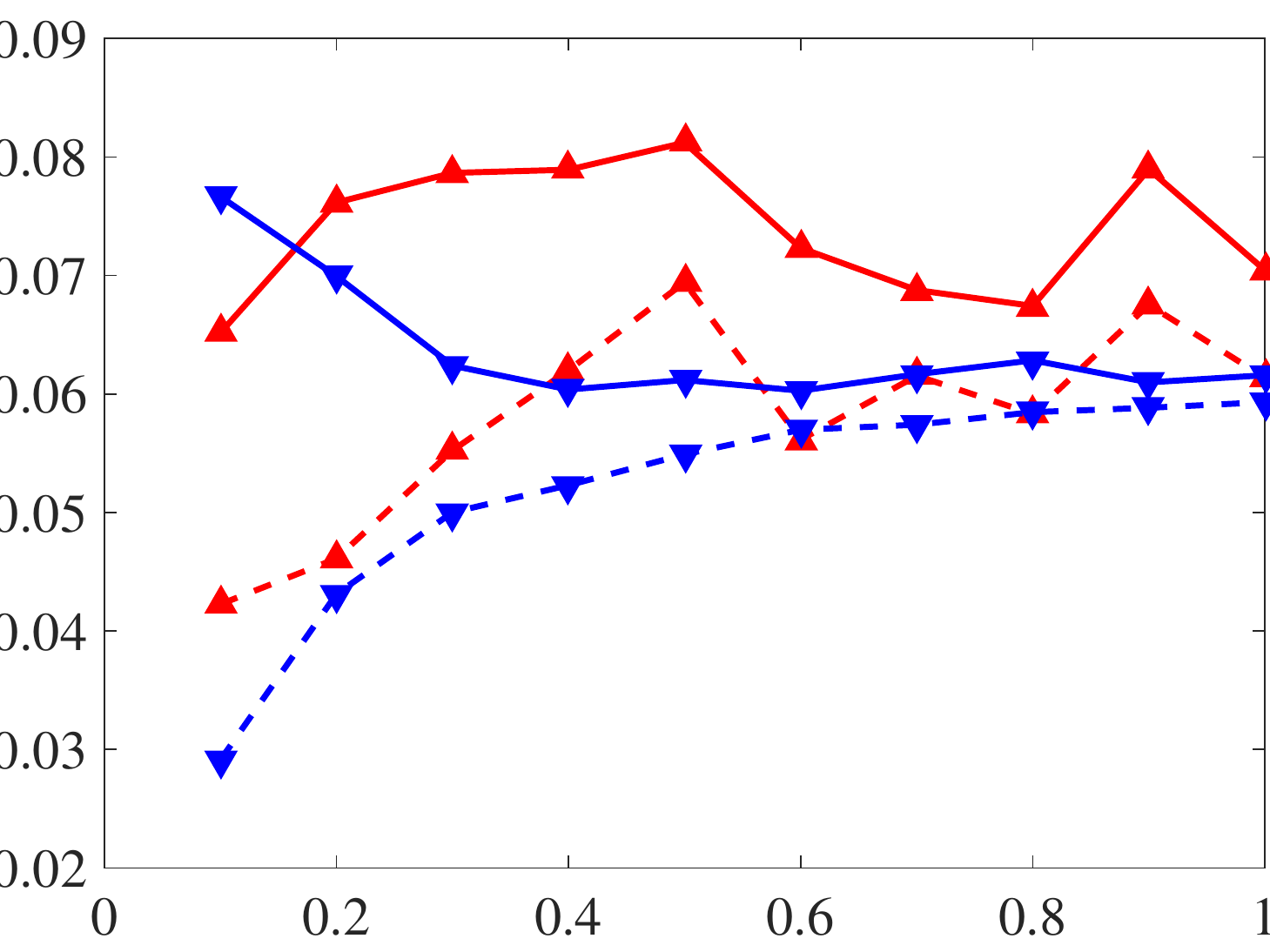} &
  \includegraphics[width=0.40\textwidth]{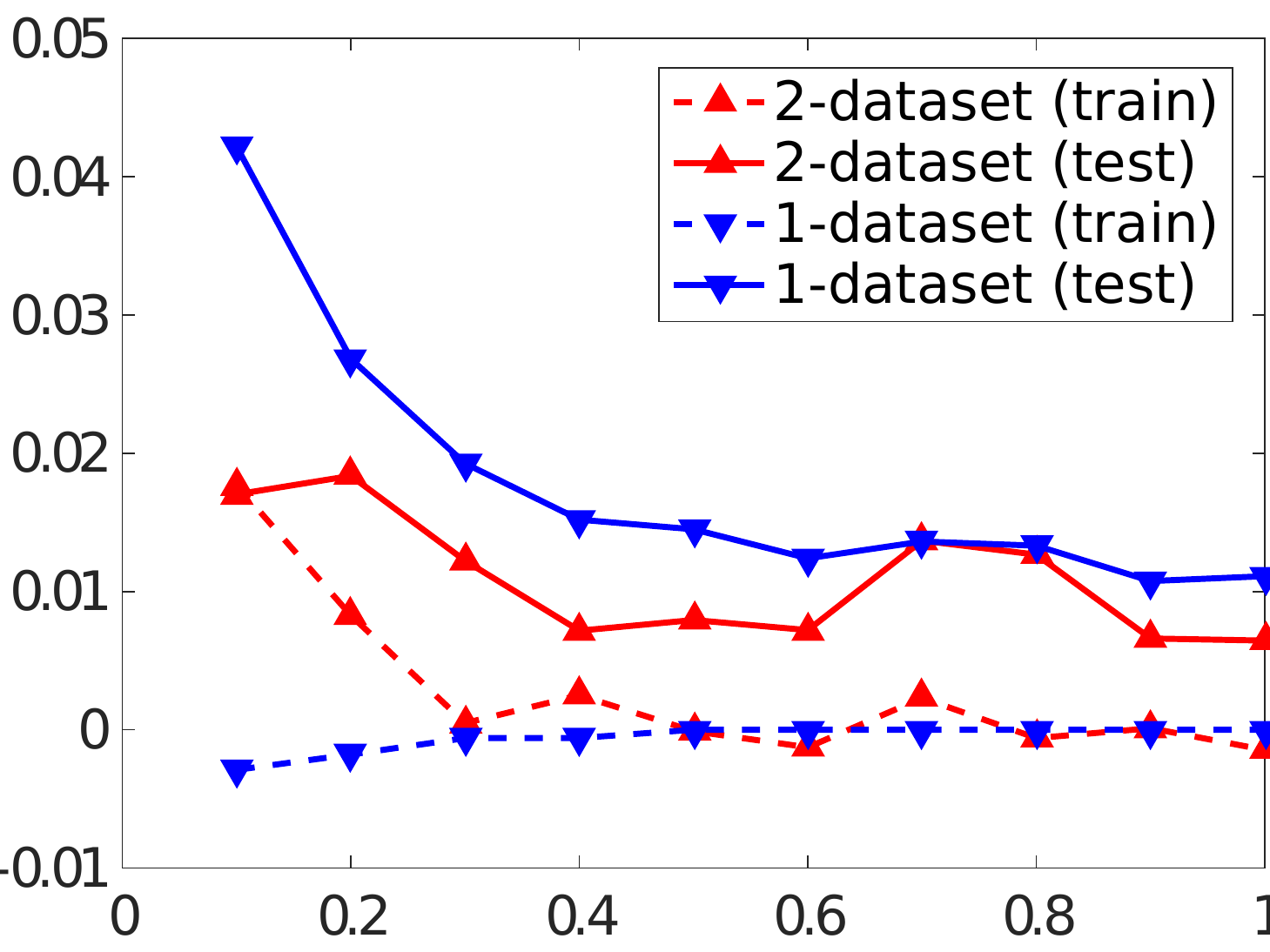} \\
  $\sigma$ & $\sigma$
\end{tabular}

\caption{
  Results of the experiment of \secref{experiments:simulated}, using
  \algref{practical} on a linear model.
  The left-hand plot demonstrates both that overfitting increases as $\sigma$
  decreases, as expected, and that our model performs worse than the baseline,
  in terms of accuracy.
  In the right-hand plot, however, we can see that our approach generalizes
  better on the constraints (the two red curves are closer together than the
  two blue ones), and also does a better job of satisfying the constraints on
  the testing set (the solid red curve is below the solid blue one).
}

\label{fig:simulated}

\end{figure*}

\begin{figure*}[t]

\centering

\begin{tabular}{cc}
  %
  %
  \multicolumn{2}{c}{\textbf{5 Hidden Units}} \\
  \textbf{0/1 error} & \textbf{Constraint violation} \\
  \includegraphics[width=0.40\textwidth]{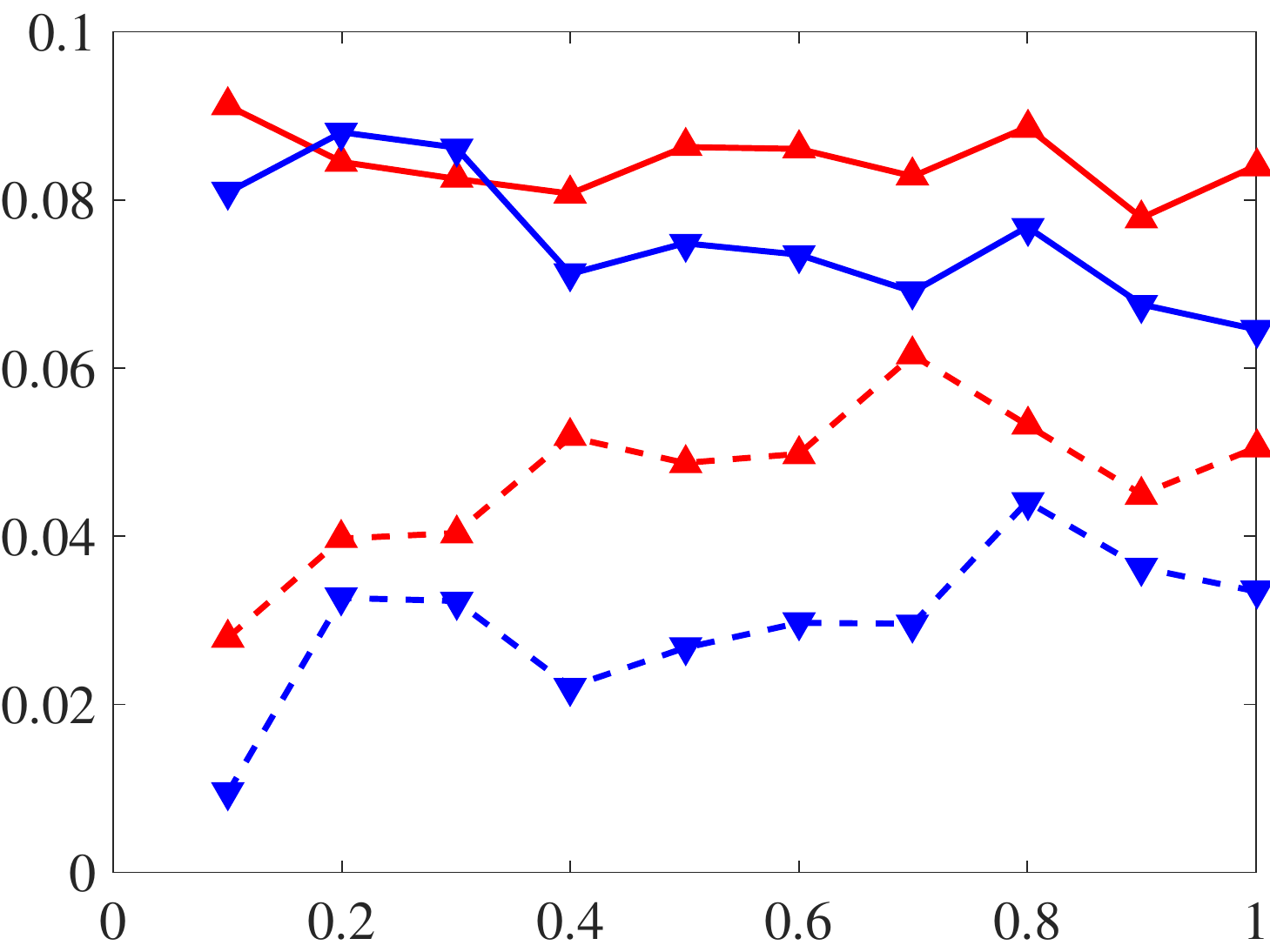} &
  \includegraphics[width=0.40\textwidth]{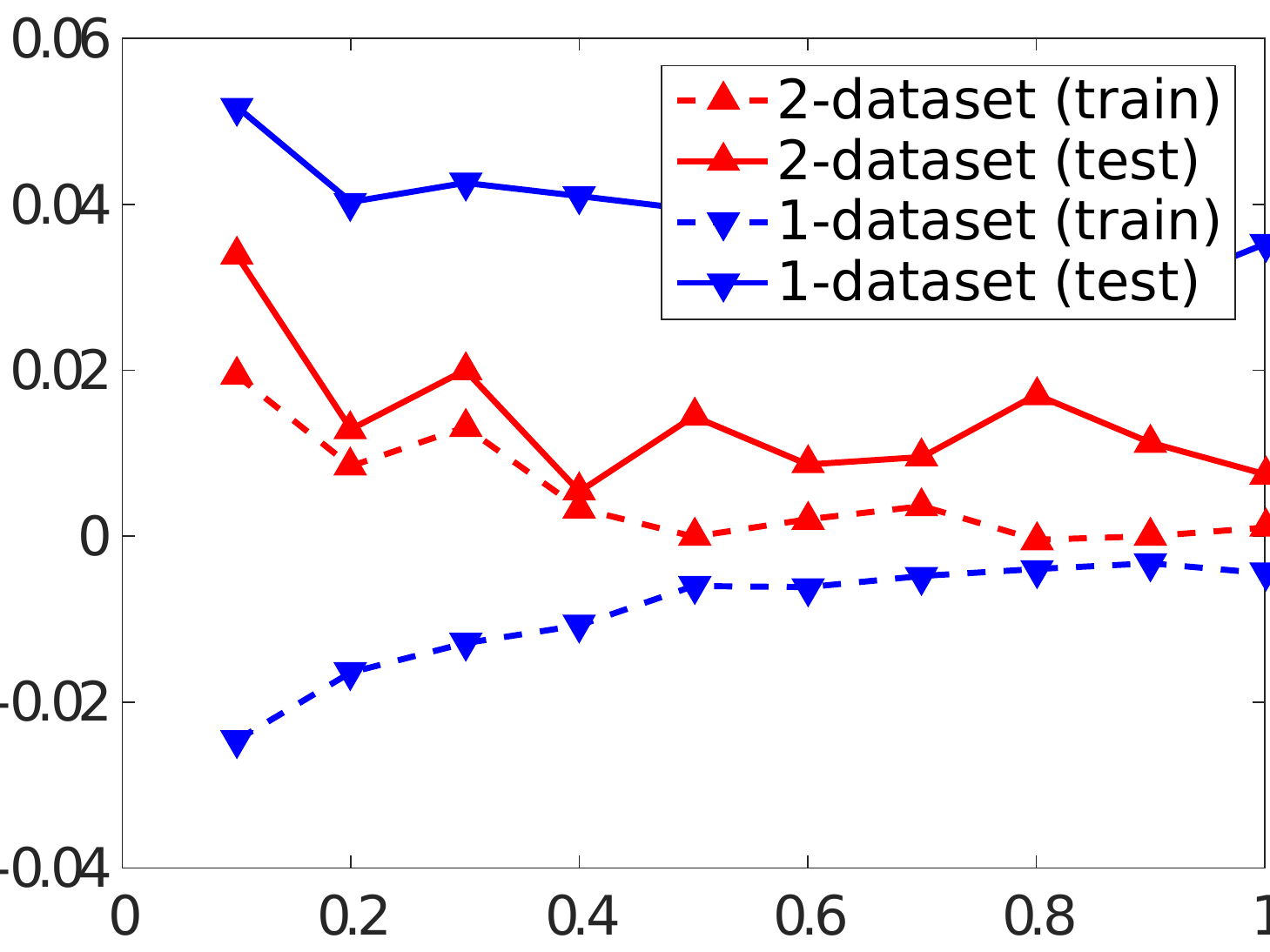} \\
  \vspace{0.5em} & \\
  \multicolumn{2}{c}{\textbf{10 Hidden Units}} \\
  \textbf{0/1 error} & \textbf{Constraint violation} \\
  \includegraphics[width=0.40\textwidth]{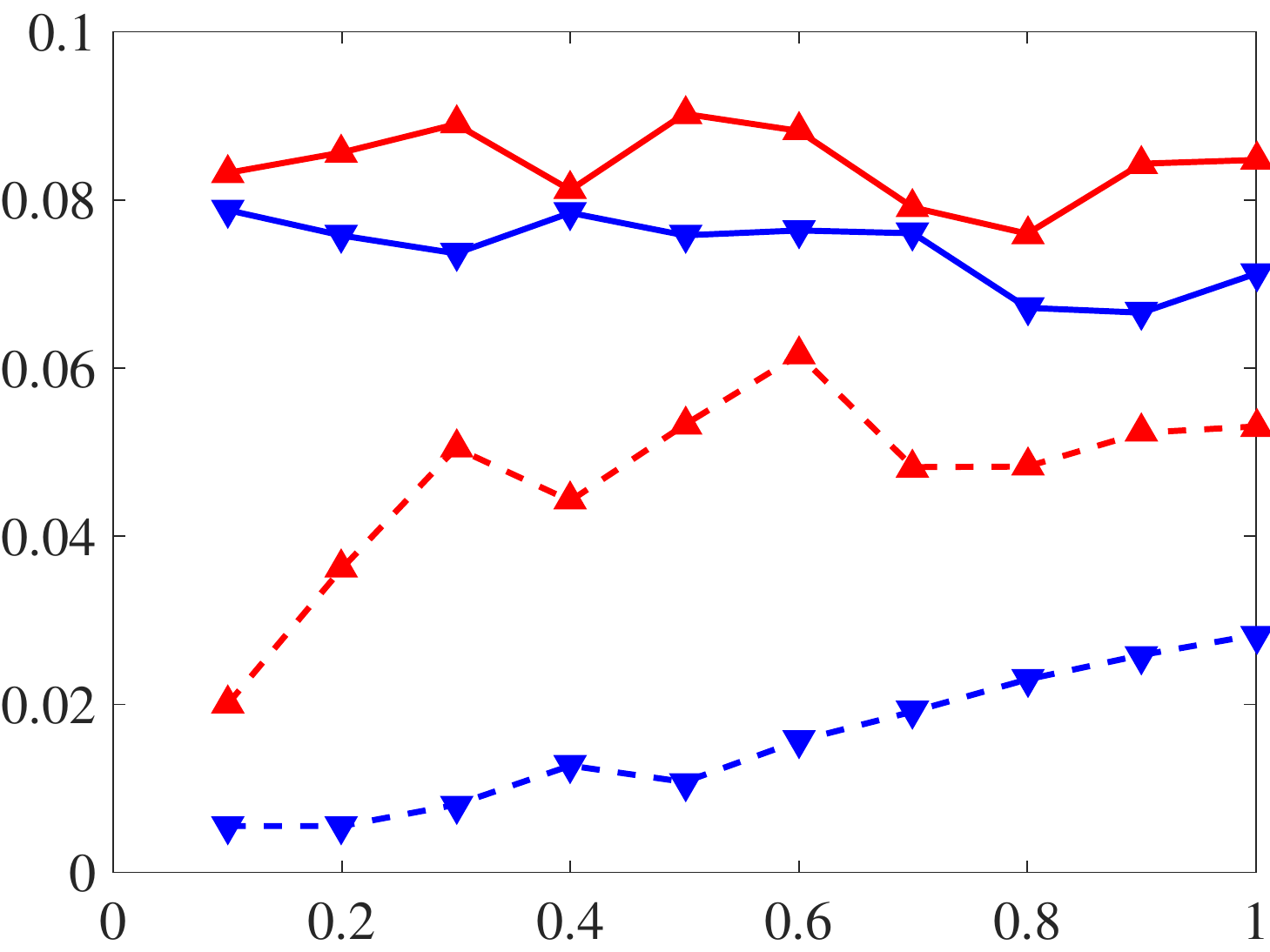} &
  \includegraphics[width=0.40\textwidth]{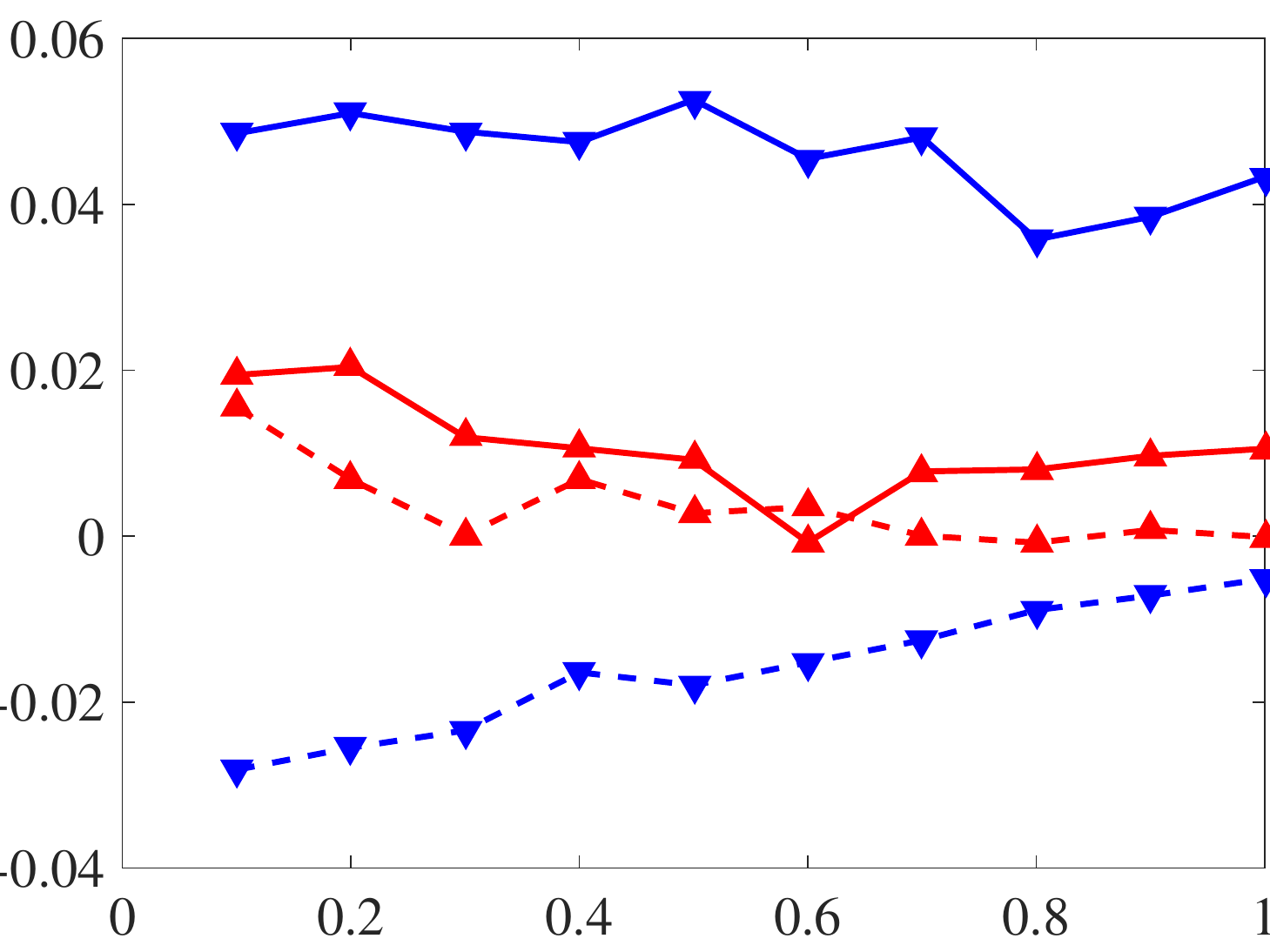} \\
  \vspace{0.5em} & \\
  \multicolumn{2}{c}{\textbf{100 Hidden Units}} \\
  \textbf{0/1 error} & \textbf{Constraint violation} \\
  \includegraphics[width=0.40\textwidth]{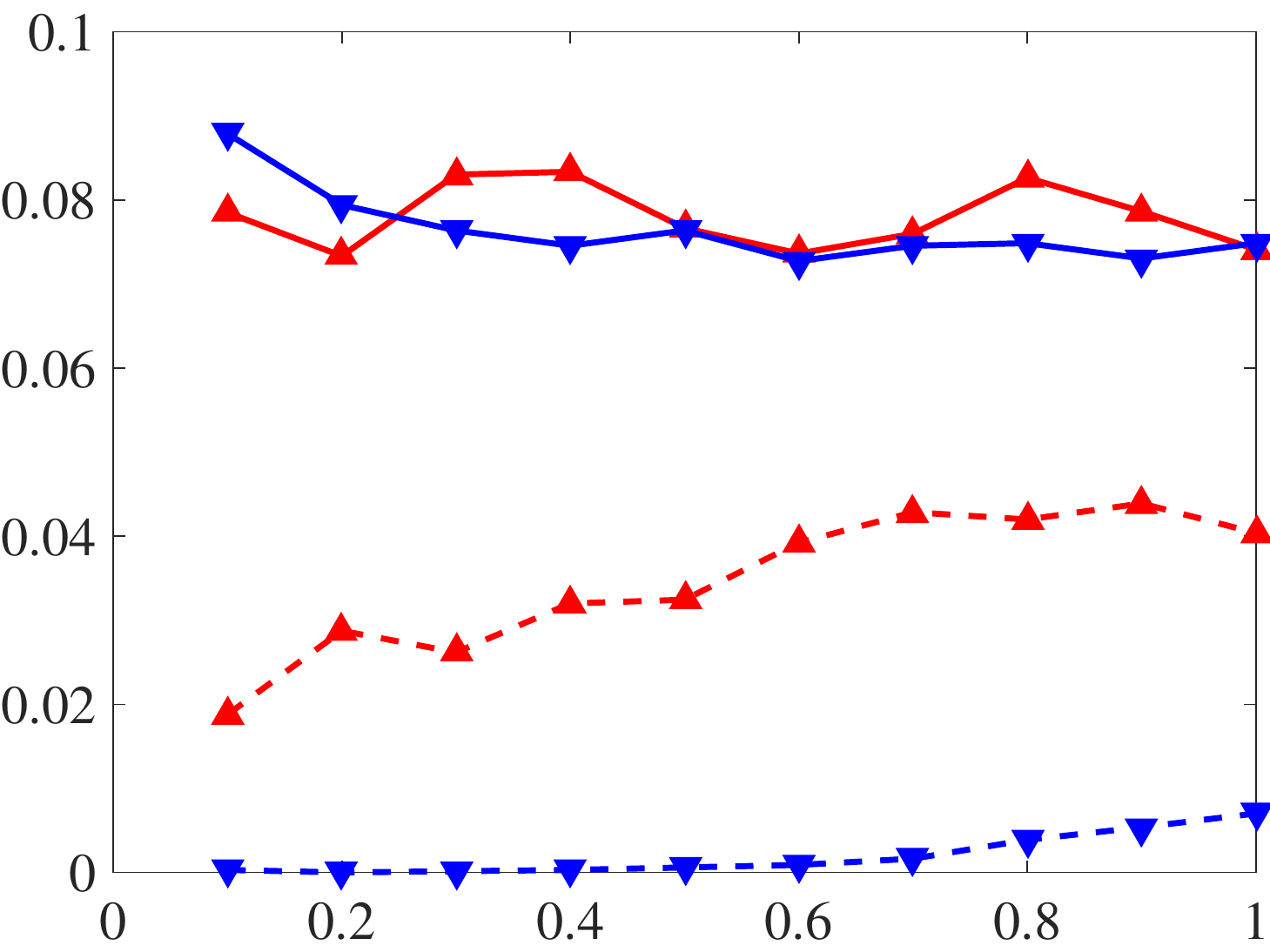} &
  \includegraphics[width=0.40\textwidth]{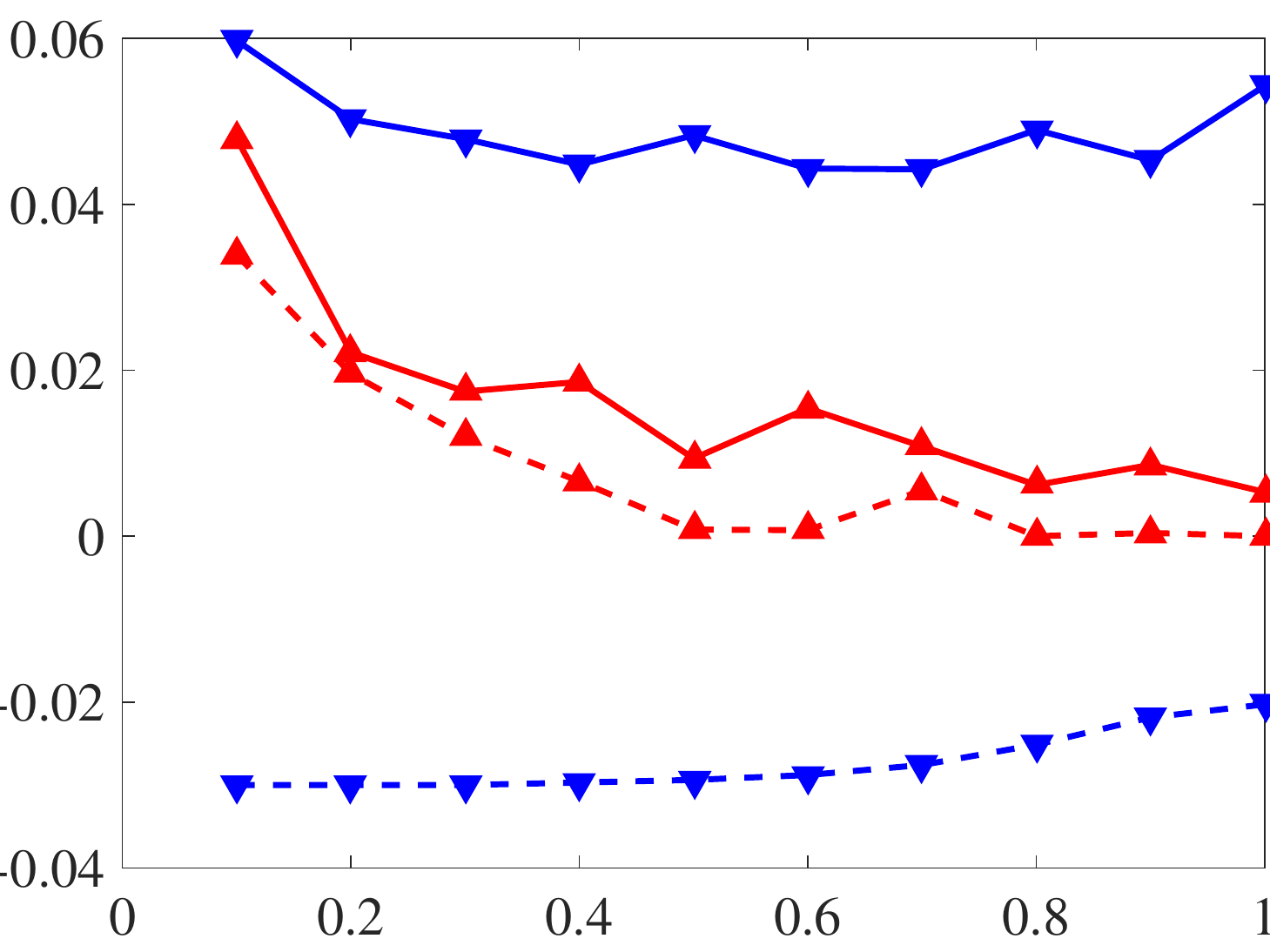} \\
  $\sigma$ & $\sigma$
\end{tabular}

\caption{
  Same as \figref{simulated}, but for one-hidden-layer neural networks with
  $5$, $10$ and $100$ hidden units. The results follow the same general trend
  as those of \figref{simulated}, but, just as one would expect, the benefits
  of our proposed two-dataset approach become more pronounced as the model
  complexity increases.
}

\label{fig:simulated-all}

\end{figure*}

Our first experiment uses a simulated binary classification problem designed to
be especially prone to overfitting. To generate the dataset, we first draw $n =
1000$ points $z_1,\dots,z_n$ from two overlapping Gaussians in $\R^2$, and
another $n$ points $w_1,\dots,w_n$ from the same distribution. For each $i$, we
let the classification label $y_i$ indicate which of the two Gaussians $z_i$
was drawn from, and generate a feature vector $x_i \in \mathcal{X} \defeq \R^n$
such that the $j$th feature satisfies $x_{i,j} \defeq \exp( -\norm{ z_i - w_j
}^2 / 2\sigma^2 )$. Our results are averaged over ten runs, with different
random splits of the data into equally-sized training, validation and testing
datasets.

The classification task is learn a classifier on $\mathcal{X}$ that determines
which of the two Gaussian distributions generated the example, with the
model's recall constrained to be at least $97\%$.
The $\sigma$ parameter partly controls the amount of overfitting: as $\sigma
\rightarrow 0$, a linear classifier on $\mathcal{X}$ approaches a 1-nearest
neighbor classifier over $w_1,\dots,w_n$, which one would expect to overfit
badly.

We trained four sets of models using \algref{practical}: linear, and
one-hidden-layer neural networks with $5$, $10$ and $100$ hidden ReLU units. We
also varied $\sigma$ between $0$ and $1$. \figrefs{simulated}{simulated-all}
show that our approach consistently comes closer to satisfying the constraints
on the testing set, but that, as one would expect, this comes at a slight cost
in testing accuracy. Unsurprisingly, our approach is most advantageous for the
most complex models ($100$-hidden unit), and less so for the simplest (linear).

\subsection{Real-data Experiments}\label{sec:experiments:real}

\begin{table*}[t]

\centering

\caption{
  Properties of datasets used in \secref{experiments:real}. For the one-dataset
  experiments, the entire training set was used for training. For the
  two-dataset experiments, the training dataset was split in half between
  $\traindataset$ and $\valdataset$.
}

\label{tab:datasets}

\begin{tabular}{r|c|ccc}
  \toprule
  \textbf{Dataset} & \textbf{Model} & \textbf{Training examples} & \textbf{Testing examples} & \textbf{Features} \\
  \midrule
  \textbf{Communities and Crime} & Linear & $1\,459$ & $499$ & $140$ \\
  \textbf{Business Entity Resolution} & Lattice & $11\,560$ & $3\,856$ & $37$ \\
  \textbf{Adult} & Neural Network & $32\,561$ & $16\,281$ & $122$ \\
  \textbf{COMPAS} & Neural Network & $4\,110$ & $2\,026$ & $32$ \\
  \bottomrule
\end{tabular}

\end{table*}

\begin{table*}[t]

\centering

\caption{
  Error rates and maximum constraint violations for all compared algorithms, on
  the datasets of \tabref{datasets}, as described in \secref{experiments:real}.
  The ``Unconstrained'' columns contain the results for entirely-unconstrained
  models.
  All quantities are averaged over $100$ runs. The training constraint
  violations are occasionally exactly zero thanks to our use of
  \citet{Cotter:2018}'s ``shrinking'' procedure (\appref{shrinking}).
}

\label{tab:real}

\begin{tabular}{c|c|cc|cc|cc|cc|cc}
\toprule
\multicolumn{2}{c|}{} & \multicolumn{2}{c|}{} & \multicolumn{4}{c|}{\textbf{\algref{practical}}} & \multicolumn{4}{c}{\textbf{\algref{lagrangian-practical}}} \\
\multicolumn{2}{c|}{} & \multicolumn{2}{c|}{\textbf{Unconstrained}} & \multicolumn{2}{c|}{One-dataset} & \multicolumn{2}{c|}{Two-dataset} & \multicolumn{2}{c|}{One-dataset} & \multicolumn{2}{c}{Two-dataset} \\
\multicolumn{2}{c|}{\textbf{Dataset}} & Error & Viol. & Error & Viol. & Error & Viol. & Error & Viol. & Error & Viol. \\
\midrule
\textbf{Communities} & Train & $.121$ & $.231$ & $.153$ & $0$ & $.161$ & $0$ & $.163$ & $-.001$ & $.165$ & $0$ \\
\textbf{and Crime} & Test & $.142$ & $.300$ & $.173$ & $.022$ & $.199$ & $-.008$ & $.181$ & $.001$ & $.195$ & $-.012$ \\
%
%
\hline
\textbf{Entity} & Train & $.148$ & $.309$ & $.216$ & $.026$ & $.215$ & $.040$ & $.225$ & $0$ & $.261$ & $.003$ \\
\textbf{Resolution} & Test & $.156$ & $.278$ & $.222$ & $.073$ & $.221$ & $.072$ & $.232$ & $.042$ & $.267$ & $.041$ \\
%
%
\hline
\multirow{2}{*}{\textbf{Adult}} & Train & $.102$ & $.077$ & $.132$ & $0$ & $.110$ & $0$ & $.131$ & $0$ & $.113$ & $0$ \\
& Test & $.156$ & $.075$ & $.156$ & $.011$ & $.169$ & $.005$ & $.156$ & $.013$ & $.165$ & $.008$ \\
%
%
\hline
\multirow{2}{*}{\textbf{COMPAS}} & Train & $.216$ & $.004$ & $.216$ & $-.005$ & $.154$ & $-.003$ & $.216$ & $-.005$ & $.151$ & $-.003$ \\
& Test & $.353$ & $.046$ & $.353$ & $.038$ & $.378$ & $.004$ & $.349$ & $.029$ & $.378$ & $.006$ \\
%
%
\bottomrule
\end{tabular}

\end{table*}

Our next set of experiments were performed on four real datasets, which we
summarize in \tabref{datasets}, and describe below. On each dataset, we trained
one of three different types of models: linear, calibrated
lattice~\citep{Gupta:2016}, or a neural network with one hidden layer
containing $50$ ReLU neurons. The neural network models are more complex than
necessary for these datasets, and are used here \emph{because} they overfit,
and therefore illustrate the improved generalization performance of the
two-dataset approach.

\begin{titled-paragraph}{Communities and Crime}
This UCI dataset~\citep{UCI} includes features aggregated from census and law
enforcement data, on which we train a linear model. The binary classification
task is to predict whether a community has a high (above the $70$th percentile)
or low crime rate, as in \citet{Kearns:2017}.
To form protected groups, we use four racial percentage features as real-valued
protected attributes. Each is thresholded at the $50$th percentile to form
eight protected groups: low-Asian, high-Asian, low-Black, high-Black,
low-Hispanic, high-Hispanic, low-White, and high-White.
There is one fairness constraint for each of the eight protected groups, which
constrains the group's false positive rate to be no larger than the overall
false positive rate.
\end{titled-paragraph}

\begin{titled-paragraph}{Business Entity Resolution}
This is a proprietary \google dataset for which the task is to predict whether
a pair of business descriptions describe the same real business. Features
include measures of similarity between the two business titles, phone numbers,
and so on. We impose several constraints: (i) for each of the $16$ most common
countries, the recall must be at least $95\%$; (ii) for the set of all chain
businesses, and likewise for the set of all non-chain businesses, the recall
must be at least $95\%$; (iii) the accuracy on non-chain businesses must be no
more than $10\%$ higher then that on chain businesses. The purpose of this
final constraint is to attempt to treat small businesses and large businesses
comparably.
\end{titled-paragraph}

\begin{titled-paragraph}{Adult}
This is a version of the UCI Adult dataset, preprocessed to include only binary
features (using one-hot encodings for categorical features, and bucketing for
continuous features). The classification task is to predict whether a person's
yearly income is greater than $\$50\,000$, subject to the $80\%$ rule for
demographic parity: for each of four overlapping protected classes (Black,
White, Female and Male), the positive prediction rate must be at least $80\%$
of the overall positive prediction rate.
\end{titled-paragraph}

\begin{titled-paragraph}{COMPAS}
This is the ProPublica COMPAS dataset analyzed by \citet{Angwin:2016},
preprocedded similarly to the Adult dataset. The classification task is to
predict recidivism, subject to equal opportunity~\citep{Hardt:2016} fairness
constraints: for each of four overlapping protected classes (Black, White,
Female and Male), the positive prediction rate on the positively-labeled
examples must be at most $5\%$ higher than the overall positive prediction rate
on positively-labeled examples.
\end{titled-paragraph}

All of these datasets have designated training/testing splits. For the
one-dataset experiments, both $\traindataset$ and $\valdataset$ were taken to
be the entire training set. For the two-dataset experiments, the training
dataset was split in half, into $\traindataset$ and $\valdataset$. All reported
numbers are averaged over $100$ such random splits, with random permutations
applied to the data.

\tabref{real} summarizes the results of these experiments. On three of the four
datasets (all but Business Entity Resolution), and for both
\algrefs{practical}{lagrangian-practical}, the two-dataset experiments have a
clear and significant advantage in terms of constraint generalization
performance, although it comes at a cost: the error rates are, as expected,
somewhat higher. While one must be mindful of this trade-off, it seems
that providing independent datasets to the $\parameters$- and
$\multipliers$-players does improve constraint generalization in practice, even
when our proofs do not apply.

\newpage
\clearpage

\bibliography{main}
\bibliographystyle{plainnat}

\newpage
\clearpage
\onecolumn

\appendix

\showproofstrue

\begin{table*}[!ht]

\centering

\caption{Key notation.}

\label{tab:notation}

\begin{tabular}{lll}
  \toprule
  \textbf{Symbol} & \textbf{Description} \\
  \midrule
  $\numconstraints$ & Number of constraints in \eqref{constrained-problem} \\
  $\objectiveloss$ & Objective loss function in \eqref{constrained-problem} \\
  $\constraintloss{i}$ & $i$th constraint loss function in \eqref{constrained-problem} \\
  $\proxyconstraintloss{i}$ & $i$th proxy-constraint loss function (corresponding to $\constraintloss{i}$) in \defref{empirical-proxy-lagrangians} \\
  $x$ & Feature vector for an example \\
  $\mathcal{X}$ & Space of feature vectors $x \in \mathcal{X}$ \\
  $\datadistribution$ & Data distribution over feature vectors $x\in\mathcal{X}$ \\
  $\traindataset$ & An \iid training sample from $\datadistribution$ \\
  $\valdataset$ & An \iid ``validation'' sample from $\datadistribution$ \\
  $\parameters$ & Parameter vector defining a model \\
  $\Parameters$ & Space of parameter vectors ($\parameters \in \Parameters$) \\
  $\outputParameters$ & Subset of $\Parameters$ consisting of the sequence of algorithm outputs $\parameters^{(1)},\dots,\parameters^{(T)}$ \\
  $\bar{\parameters}$ & Random variable over $\outputParameters$ defining a stochastic model (\thmrefs{discrete}{continuous}) \\
  $\multipliers$ & Vector of Lagrange-multiplier-like ``hyperparameters'' (\defref{empirical-proxy-lagrangians}) \\
  $\Multipliers$ & Space of Lagrange-multiplier-like ``hyperparameter'' vectors ($\multipliers \in \Multipliers \defeq \Delta^{\numconstraints + 1}$) \\
  $\bar{\multipliers}$ & Average ``hyperparameters'' $\bar{\multipliers} \defeq ( \sum_{t=1}^T \multipliers^{(t)} ) / T$ (\thmrefs{discrete}{continuous}) \\
  $\bar{\multipliers}_1$ & First coordinate of $\bar{\multipliers}$, measuring our ``belief'' that $\bar{\parameters}$ is feasible \\
  $\lagrangian_{\parameters}$ & In-expectation proxy-Lagrangian function minimized by $\parameters$-player (\defref{proxy-lagrangians}) \\
  $\lagrangian_{\multipliers}$ & In-expectation proxy-Lagrangian function maximized by $\multipliers$-player (\defref{proxy-lagrangians}) \\
  $\empiricallagrangian_{\parameters}$ & Empirical proxy-Lagrangian function minimized by $\parameters$-player (\defref{empirical-proxy-lagrangians}) \\
  $\empiricallagrangian_{\multipliers}$ & Empirical proxy-Lagrangian function maximized by $\multipliers$-player (\defref{empirical-proxy-lagrangians}) \\
  $\traingeneralization{\Parameters}$ & Generalization bound for $\objectiveloss,\proxyconstraintloss{1},\dots,\proxyconstraintloss{\numconstraints}$ on $\traindataset$ for $\parameters \in \Parameters$ (\defref{generalization}) \\
  $\valgeneralization{\outputParameters}$ & Generalization bound for $\constraintloss{1},\dots,\constraintloss{\numconstraints}$ on $\valdataset$ for $\parameters \in \outputParameters$ (\defref{generalization}) \\
  $\oracle$ & Bayesian oracle of \defref{oracle} \\
  $\approximation$ & Additive approximation of the Bayesian oracle of \defref{oracle} \\
  $\covering$ & A radius-$\coveringradius$ external covering of $\Multipliers \defeq \Delta^{\numconstraints+1}$ \wrt the $1$-norm \\
  $\coveringradius$ & Radius of the covering $\covering$ \\
  $\coveringParameters$ & Oracle evaluations at the covering centers ($\coveringParameters \defeq \{ \oracle( \empiricallagrangian_{\parameters}( \cdot, \tilde{\multipliers} )) : \tilde{\multipliers} \in \covering \}$) \\
  $T$ & Number of iterations performed in \algrefss{discrete}{practical}{lagrangian-practical} \\
  $T_{\multipliers}$ & Number of iterations performed in the outer loop of \algref{continuous} \\
  $T_{\parameters}$ & Number of iterations performed in the inner loop of \algref{continuous} \\
  $\eta_{\multipliers}$ & Step size associated with $\multipliers$-player in \algrefsss{discrete}{continuous}{practical}{lagrangian-practical} \\
  $\eta_{\parameters}$ & Step size associated with $\parameters$-player in \algrefs{practical}{lagrangian-practical} \\
  $\matrixmultipliers$ & A left-stochastic $(\numconstraints+1)\times(\numconstraints+1)$ matrix \\
  $\Matrixmultipliers$ & Space of all left-stochastic $(\numconstraints+1)\times(\numconstraints+1)$ matrices ($\matrixmultipliers \in \Matrixmultipliers$) \\
  $\margin$ & Maximum margin by which the proxy-constraints can be satisfied in \thmrefs{discrete}{continuous} \\
  $\strongconvexity$ & Strong convexity parameter of $\objectiveloss, \proxyconstraintloss{1}, \dots, \proxyconstraintloss{\numconstraints}$ in \thmref{continuous} \\
  $\lipschitz$ & Lipschitz constant of $\constraintloss{1}, \dots, \constraintloss{\numconstraints}$ in \thmref{continuous} \\
  $\bound{\objectiveloss}$ & Upper bound on $b_0 - a_0$, where $\range( \objectiveloss ) = [a_0, b_0]$ \\
  $\bound{\loss}$ & Upper bound on $b_i - a_i$ for all $i \in \indices{\numconstraints}$, where $\range( \constraintloss{i} ) = [a_i, b_i]$ \\
  $\bound{\tilde{\loss}}$ & Upper bound on $\abs{ \ell(x;\parameters) }$ for all $\ell \in \{ \objectiveloss, \proxyconstraintloss{1}, \dots, \proxyconstraintloss{\numconstraints} \}$ \\
  $\bound{\stochasticsubgrad}$ & Upper bound on the $2$-norms of subgradients of $\empiricallagrangian_{\parameters}$ \wrt $\parameters$ \\
  $\bound{\stochasticgrad}$ & Upper bound on the $\infty$-norms of gradients of $\empiricallagrangian_{\multipliers}$ \wrt $\multipliers$ \\
  \bottomrule
\end{tabular}

\end{table*}

\section{Examples of Constraints}\label{app:examples}

In this appendix, we'll provide some examples of constrained optimization
problems in the form of \eqref{constrained-problem}.

\subsection{Neyman-Pearson}\label{sec:examples:neyman-pearson}

The first example we'll consider is Neyman-Pearson
classification~\citep{Davenport:2010,Bottou:2011}.  Imagine that we wish to
learn a classification function $f:\mathcal{X}\times\Parameters \rightarrow \R$
parameterized by $\parameters$, with the goal being to minimize the false
positive rate, subject to the constraint that the false negative rate be at
most $10\%$:
\begin{align*}
  \minimize[\parameters \in \Parameters] & \expectation_{x,y \mid y = -1}\left[
  \indicator\left\{ f\left(x; \parameters\right) \ge 0 \right\} \right] \\
  \suchthat & \expectation_{x,y \mid y = 1}\left[ \indicator\left\{ f\left(x;
  \parameters\right) \le 0 \right\} \right] \le 0.1
\end{align*}
One way to convert this problem into the form of \eqref{constrained-problem} is
to define $\datadistribution_{+}$ and $\datadistribution_{-}$ as the marginal
distributions over $x$s for which $y=+1$ and $y=-1$ (respectively), and take
$\datadistribution \defeq \datadistribution_{+} \times \datadistribution_{-}$
so that \datadistribution is a distribution over \emph{pairs} of feature
vectors, the first having a positive label, and the second a negative label.
Defining $\objectiveloss\left(x_{+},x_{-}; \parameters\right) \defeq
\indicator\left\{ f\left(x_{-}\right) \ge 0 \right\}$ and
$\constraintloss{1}\left(x_{+},x_{-}; \parameters\right) \defeq
\indicator\left\{ f\left(x_{+}\right) \le 0 \right\} - 0.1$ puts the original
Neyman-Pearson problem in the form of \eqref{constrained-problem}.

In practice, the fact that $\objectiveloss$ and $\constraintloss{1}$ are
defined in terms of indicator functions, and are therefore discontinuous, will
problematic. To fix this, using the formulation of
\defref{empirical-proxy-lagrangians}, one could instead define
$\objectiveloss\left(x_{+},x_{-}; \parameters\right) \defeq \max\left\{ 0, 1 +
f\left(x_{-}\right) \right\}$ as a hinge upper bound on the false positive
rate, and leave $\constraintloss{1}$ as-is while defining the corresponding
proxy-constraint to be $\proxyconstraintloss{1}\left(x_{+},x_{-};
\parameters\right) \defeq \max\left\{ 0, 1 - f\left(x_{+}\right) \right\}$.

\subsection{Equal Opportunity}\label{sec:examples:equal-opportunity}

The second example we'll consider is a fairness-constrained problem. As before,
we'll take $f:\mathcal{X}\times\Parameters \rightarrow \R$ to be a
classification function, but we'll imagine that each $x\in\mathcal{X}$ contains
a feature $x_k \in \left\{1,2,3\right\}$ indicating to which of three protected
classes the corresponding example belongs. We will seek to minimize the overall
error rate, subject to the constraint that, for each of the three protected
classes, the false negative rate is at most $110\%$ of the false negative rate
across all three classes (this is essentially an equal opportunity
constraint~\citep{Hardt:2016}):
\begin{align*}
  \minimize[\parameters \in \Parameters] & \expectation_{x,y}\left[
  \indicator\left\{ y f\left(x; \parameters\right) \le 0 \right\} \right] \\
  \suchthat[\forall i \in \left\{1,2,3\right\}] & \expectation_{x,y \mid y = 1
  \wedge x_k = i}\left[ \indicator\left\{ f\left(x; \parameters\right) \le 0
  \right\} \right] \le 1.1 \cdot \expectation_{x,y \mid y = 1}\left[
  \indicator\left\{ f\left(x; \parameters\right) \le 0 \right\} \right]
\end{align*}
While we could use the same approach as in the Neyman-Pearson example, \ie
taking marginals and crossing them to define a data distribution over tuples of
examples, we'll instead take $\mathcal{D}$ to be the data distribution over
$\mathcal{X}\times\left\{\pm 1\right\}$ pairs, and use the indicator feature
$x_k$ to define:
\begin{align*}
  \objectiveloss\left(x,y;\parameters\right) \defeq & \indicator\left\{ y
  f\left(x; \parameters\right) \le 0 \right\} \\
  \constraintloss{i}\left(x,y;\parameters\right) \defeq & \frac{\indicator\{ y
  = 1 \wedge x_k = i \} \indicator\left\{ f\left(x; \parameters\right) \le 0
  \right\}}{ \probability\{ y = 1 \wedge x_k = i \mid x,y \sim
  \datadistribution \} } - 1.1 \cdot \frac{\indicator\{ y = 1 \}
  \indicator\left\{ f\left(x; \parameters\right) \le 0 \right\}}{
  \probability\{ y = 1 \mid x,y \sim \datadistribution \} }
\end{align*}
for all $i \in \left\{1,2,3\right\}$, where we assume that the probabilities in
the denominators of the ratios defining $\constraintloss{i}$ are constants
known a priori.

As in the Neyman-Pearson example, in practice the indicators in the objective
function could be replaced with differentiable upper bounds, and a
differentiable proxy-constraint $\proxyconstraintloss{i}$ could be introduced
for each $\constraintloss{i}$.

\section{Shrinking}\label{app:shrinking}

\citet{Cotter:2018} introduced a procedure for ``shrinking'' the support size
of a $\bar{\parameters}$. When adapted to our setting, the first step is to
evaluate the objective and constraints for every iterate (in practice, this
is overkill; one should subsample the iterates):
\begin{align*}
  \vec{\objectiveloss}^{(t)} =& \frac{1}{\abs{\traindataset}} \sum_{x \in
  \traindataset} \objectiveloss \left( x; \parameters^{(t)}\right) \\
  \vec{\constraintloss{i}}^{(t)} =& \frac{1}{\abs{\valdataset}} \sum_{x \in
  \valdataset} \constraintloss{i}\left(x; \parameters^{(t)}\right)
\end{align*}
Next, we optimize a linear program (LP) that seeks a distribution $p$ over
$\outputParameters$ that minimizes the objective while violating no constraint
by more than $\epsilon$:
\begin{equation*}
  \min_{p \in \Delta^T} \inner{p}{\vec{\objectiveloss}}
  \;\;\;\;
  \suchthat[\forall i \in \indices{\numconstraints}]
  \inner{p}{\vec{\constraintloss{i}}} \le \epsilon
\end{equation*}
Notice that the $p \in \Delta^{T}$ condition adds several implicit simplex
constraints to this LP. The key to this procedure is that, as
\citet{Cotter:2018} show, every vertex $p$ of this linear program has at most
$\numconstraints+1$ nonzero elements, where $\numconstraints$ is the number of
constraints.

In particular, if $\epsilon$ is chosen to be the maximum validation constraint
violation of the ``original'' stochastic classifier $\bar{\parameters}$, then
an \emph{optimal} vertex $p^*$ will have a training objective function value
and maximum validation constraint violation that are no larger than those of
$\bar{\parameters}$, and $p^*$ will be supported on only $\numconstraints+1$
$\parameters^{(t)}$s.
Furthermore, since the resulting stochastic classifier is still supported on a
subset of $\outputParameters$, the generalization definitions of
\secref{background:generalization} apply to it just as well as they did to
$\bar{\parameters}$.

While it would be possible to provide optimality and feasibility guarantees for
the result of this ``shrinking'' procedure, we will only use it in our
experiments (\secref{experiments}). There, instead of taking the LP's
$\epsilon$ parameter to be the maximum validation constraint violation of
$\bar{\parameters}$, we follow \citet{Cotter:2018}'s suggestion to use an outer
bisection search to find the smallest $\epsilon \ge 0$ for which the LP is
feasible.

\section{Proofs}\label{app:proofs}

We'll begin by reproducing a definition from \citet{Cotter:2018}:
\begin{definition}
  \textbf{(Definition 2 of \citet{Cotter:2018})}
  \label{def:proxy-lagrangians}
  Given proxy loss functions $\proxyconstraintloss{i}\left(x;
  \parameters\right) \ge \constraintloss{i}\left(x; \parameters\right)$ for all
  $x \in \mathcal{X}$ and $i \in \indices{\numconstraints}$, the
  proxy-Lagrangians $\lagrangian_{\parameters},\lagrangian_{\multipliers} :
  \Parameters \times \Multipliers \rightarrow \R$ of
  \eqref{constrained-problem} are:
  \begin{align*}
    \lagrangian_{\parameters}\left(\parameters, \multipliers\right) \defeq&
    \expectation_{x \sim \datadistribution}\left[ \multipliers_1
    \objectiveloss\left(x; \parameters\right) + \sum_{i=1}^{\numconstraints}
    \multipliers_{i+1} \proxyconstraintloss{i}\left(x; \parameters\right)
    \right] \\
    \lagrangian_{\multipliers}\left(\parameters, \multipliers\right) \defeq&
    \expectation_{x \sim \datadistribution}\left[ \sum_{i=1}^{\numconstraints}
    \multipliers_{i+1} \constraintloss{i}\left(x; \parameters\right) \right]
  \end{align*}
  where $\Multipliers \defeq \Delta^{\numconstraints+1}$ is the
  $\left(\numconstraints+1\right)$-dimensional simplex.
\end{definition}
This definition differs from \defref{empirical-proxy-lagrangians} in that the
former is the in-expectation version of the latter, which is written in terms
of separate \iid training and validation sets.

Theorem 2 of \citet{Cotter:2018} characterizes the optimality and feasibility
properties of a particular type of $\Phi$-correlated equilibrium of
\defref{proxy-lagrangians}. We adapt it to our setting, giving the analogous
result for such an equilibrium of \defref{empirical-proxy-lagrangians}:
\begin{thm}{dataset-suboptimality}
  Define $\Matrixmultipliers$
  as the set of all left-stochastic $\left(\numconstraints + 1\right) \times
  \left(\numconstraints + 1\right)$ matrices, $\Multipliers \defeq
  \Delta^{\numconstraints+1}$ as the
  $\left(\numconstraints+1\right)$-dimensional simplex, and assume that each
  $\proxyconstraintloss{i}$ upper bounds the corresponding $\constraintloss{i}$.
  Let $\parameters^{(1)},\dots,\parameters^{(T)} \in \Parameters$ and
  $\multipliers^{(1)},\dots,\multipliers^{(T)} \in \Multipliers$ be sequences
  satisfying:
  \begin{align}
    \label{eq:dataset-suboptimality:equilibrium}
    \frac{1}{T} \sum_{t=1}^T \empiricallagrangian_{\parameters}\left(
    \parameters^{(t)}, \multipliers^{(t)} \right) - \inf_{\parameters^* \in
    \Parameters} \frac{1}{T} \sum_{t=1}^T
    \empiricallagrangian_{\parameters}\left( \parameters^*, \multipliers^{(t)}
    \right) \le& \epsilon_{\parameters} \\
    \notag \max_{\matrixmultipliers^* \in \Matrixmultipliers} \frac{1}{T} \sum_{t=1}^T
    \empiricallagrangian_{\multipliers}\left( \parameters^{(t)},
    \matrixmultipliers^* \multipliers^{(t)} \right) - \frac{1}{T} \sum_{t=1}^T
    \empiricallagrangian_{\multipliers}\left( \parameters^{(t)},
    \multipliers^{(t)} \right) \le& \epsilon_{\multipliers}
  \end{align}
  Define $\outputParameters \defeq \left\{ \parameters^{(1)}, \dots,
  \parameters^{(T)} \right\}$. Let $\bar{\parameters}$ be a random variable
  taking values from $\outputParameters$, defined such that $\bar{\parameters} =
  \parameters^{(t)}$ with probability $\multipliers^{(t)}_1 / \sum_{s=1}^T
  \multipliers^{(s)}_1$, and let $\bar{\multipliers} \defeq \left(\sum_{t=1}^T
  \multipliers^{(t)}\right) / T$.
  Then $\bar{\parameters}$ is nearly-optimal in expectation:
  \begin{align}
    \label{eq:dataset-suboptimality:optimality}
    \expectation_{\bar{\parameters}, x \sim \datadistribution}\left[
    \objectiveloss\left(x; \bar{\parameters}\right) \right] \le &
    \inf_{\parameters^* \in \Parameters : \forall i .  \expectation_{x \sim
    \datadistribution}\left[ \proxyconstraintloss{i}\left(x;
    \parameters^*\right) \right] \le 0} \expectation_{x \sim
    \datadistribution}\left[ \objectiveloss\left(x; \parameters^* \right)
    \right] \\
    \notag & + \frac{1}{\bar{\multipliers}_1} \left( \epsilon_{\parameters} +
    \epsilon_{\multipliers} + 2 \traingeneralization{\Parameters}
    + \valgeneralization{\outputParameters} \right)
  \end{align}
  and nearly-feasible:
  \begin{equation}
    \label{eq:dataset-suboptimality:feasibility}
    \max_{i \in \indices{\numconstraints}} \expectation_{\bar{\parameters}, x
    \sim \datadistribution}\left[ \constraintloss{i}\left(x;
    \bar{\parameters}\right) \right] \le
    \frac{\epsilon_{\multipliers}}{\bar{\multipliers}_1} +
    \valgeneralization{\outputParameters}
  \end{equation}
  Additionally, if there exists a $\parameters' \in \Parameters$ that satisfies
  all of the constraints with margin $\margin$ (\ie $\expectation_{x \sim
  \datadistribution}\left[ \constraintloss{i}\left(x; \parameters'\right)
  \right] \le -\margin$ for all $i \in \indices{\numconstraints}$), then:
  \begin{equation}
    \label{eq:dataset-suboptimality:lambda-bound}
    \bar{\multipliers}_1 \ge \frac{ \margin - \epsilon_{\parameters} -
    \epsilon_{\multipliers} - 2 \traingeneralization{\Parameters} -
    \valgeneralization{\outputParameters} }{\margin + \bound{\objectiveloss}}
  \end{equation}
  where $\bound{\objectiveloss} \ge \sup_{\parameters \in \Parameters}
  \expectation_{x \sim \datadistribution}\left[ \objectiveloss\left(x;
  \parameters\right) \right] - \inf_{\parameters \in \Parameters}
  \expectation_{x \sim \datadistribution}\left[ \objectiveloss\left(x;
  \parameters\right) \right]$ is a bound on the range of the objective loss.
\end{thm}
\begin{prf}{dataset-suboptimality}
  This proof closely follows those of Theorem 4 and Lemma 7 in
  \citet{Cotter:2018}---the only addition is that, in this proof, we use the
  empirical proxy-Lagrangian formulation with separate training and validation
  datasets (\defref{empirical-proxy-lagrangians}), instead of the
  in-expectation proxy-Lagrangian (\defref{proxy-lagrangians}), and therefore
  need to account for generalization.

  \begin{titled-paragraph}{Optimality}
    If we choose $M^*$ to be the matrix with its first row being all-one, and
    all other rows being all-zero, then
    $\empiricallagrangian_{\multipliers}\left( \parameters,
    \matrixmultipliers^* \multipliers \right) = 0$, which shows that the first
    term in the LHS of the second line of
    \eqref{dataset-suboptimality:equilibrium} is nonnegative. Hence:
    \begin{equation*}
      -\expectation_{t \sim \indices{T}} \left[
      \empiricallagrangian_{\multipliers}\left(\parameters^{(t)},
      \multipliers^{(t)}\right) \right] \le \epsilon_{\multipliers}
    \end{equation*}
    so by the definitions of $\empiricallagrangian_{\multipliers}$
    (\defref{empirical-proxy-lagrangians}) and
    $\valgeneralization{\outputParameters}$ (\defref{generalization}), and the
    facts that $\proxyconstraintloss{i} \ge \constraintloss{i}$ and
    $\multipliers^{(t)} \in \Delta^{\numconstraints+1}$:
    \begin{equation*}
      \expectation_{t \sim \indices{T}, x \sim \datadistribution}\left[
      \sum_{i=1}^{\numconstraints} \multipliers^{(t)}_{i+1}
      \proxyconstraintloss{i}\left(x; \parameters^{(t)}\right) \right] \ge
      -\epsilon_{\multipliers} - \valgeneralization{\outputParameters}
    \end{equation*}
    Notice that $\empiricallagrangian_{\parameters}$ is linear in
    $\multipliers$, so the first line of
    \eqref{dataset-suboptimality:equilibrium}, combined with the definition of
    $\traingeneralization{\Parameters}$, becomes:
    \begin{align*}
      \MoveEqLeft \expectation_{t \sim \indices{T}, x \sim
      \datadistribution}\left[ \multipliers^{(t)}_1 \objectiveloss\left( x;
      \parameters^{(t)} \right) + \sum_{i=1}^{\numconstraints}
      \multipliers^{(t)}_{i+1} \proxyconstraintloss{i}\left( x;
      \parameters^{(t)} \right) \right] - \inf_{\parameters^* \in \Parameters}
      \expectation_{x \sim \datadistribution}\left[ \bar{\multipliers}_1
      \objectiveloss\left( x; \parameters^* \right) +
      \sum_{i=1}^{\numconstraints} \bar{\multipliers}_{i+1}
      \proxyconstraintloss{i}\left( x; \parameters^* \right) \right] \\
      \le& \epsilon_{\parameters} + 2 \traingeneralization{\Parameters}
    \end{align*}
    Combining the above two results:
    \begin{align}
      \label{eq:dataset-suboptimality:constraint-lower-bound}
      \MoveEqLeft \expectation_{t \sim \indices{T}, x \sim
      \datadistribution}\left[ \multipliers^{(t)}_1 \objectiveloss\left( x;
      \parameters^{(t)} \right) \right] - \inf_{\parameters^* \in \Parameters}
      \expectation_{x \sim \datadistribution}\left[ \bar{\multipliers}_1
      \objectiveloss\left( x; \parameters^* \right) +
      \sum_{i=1}^{\numconstraints} \bar{\multipliers}_{i+1}
      \proxyconstraintloss{i}\left( x; \parameters^* \right) \right] \\
      \notag \le& \epsilon_{\parameters} + \epsilon_{\multipliers} + 2
      \traingeneralization{\Parameters} + \valgeneralization{\outputParameters}
    \end{align}
    Choose $\parameters^*$ to be the optimal solution that satisfies the
    \emph{proxy} constraints, so that $\expectation_{x \sim
    \datadistribution}\left[ \proxyconstraintloss{i}\left(x;
    \parameters^*\right) \right] \le 0$ for all
    $i\in\indices{\numconstraints}$. Then:
    \begin{equation*}
      \expectation_{t \sim \indices{T}, x \sim \datadistribution}\left[
      \multipliers^{(t)}_1 \objectiveloss\left( x; \parameters^{(t)} \right)
      \right] - \expectation_{x \sim \datadistribution}\left[
      \bar{\multipliers}_1 \objectiveloss\left( x; \parameters^* \right)
      \right]
      \le \epsilon_{\parameters} + \epsilon_{\multipliers} + 2
      \traingeneralization{\Parameters} + \valgeneralization{\outputParameters}
    \end{equation*}
    which is the optimality claim.
  \end{titled-paragraph}

  \begin{titled-paragraph}{Feasibility}
    %
    %
    We'll begin by simplifying our notation: define
    $g_1\left(\parameters\right) \defeq 0$ and $g_{i+1}\left(\parameters\right)
    \defeq \expectation_{x \sim \valdataset}\left[ \constraintloss{i}\left(x;
    \parameters\right) \right]$ for $i\in\indices{\numconstraints}$, so that
    $\empiricallagrangian_{\multipliers}\left(\parameters, \multipliers\right)
    = \inner{\multipliers}{g_{:}\left(\parameters\right)}$.
    Consider the first term in the LHS of the second line of
    \eqref{dataset-suboptimality:equilibrium}:
    \begin{align*}
      \max_{\matrixmultipliers^* \in \Matrixmultipliers} \expectation_{t \sim
      \indices{T}}\left[ \empiricallagrangian_{\multipliers}\left(
      \parameters^{(t)}, \matrixmultipliers^* \multipliers^{(t)} \right)
      \right] =&
      \max_{\matrixmultipliers^* \in \Matrixmultipliers} \expectation_{t \sim
      \indices{T}}\left[ \inner{\matrixmultipliers^*
      \multipliers^{(t)}}{g_{:}\left(\parameters^{(t)}\right)} \right] \\
      =& \max_{\matrixmultipliers^* \in \Matrixmultipliers} \expectation_{t
      \sim \indices{T}}\left[ \sum_{i=1}^{\numconstraints+1}
      \sum_{j=1}^{\numconstraints+1} \matrixmultipliers^*_{j,i}
      \multipliers^{(t)}_i g_j\left(\parameters^{(t)}\right) \right] \\
      =& \sum_{i=1}^{\numconstraints+1} \max_{\matrixmultipliers^*_{:,i} \in
      \Delta^{\numconstraints+1}} \sum_{j=1}^{\numconstraints+1}
      \expectation_{t \sim \indices{T}}\left[ \matrixmultipliers^*_{j,i}
      \multipliers^{(t)}_i g_j\left(\parameters^{(t)}\right) \right] \\
      =& \sum_{i=1}^{\numconstraints+1} \max_{j \in
      \indices{\numconstraints+1}} \expectation_{t \sim \indices{T}}\left[
      \multipliers^{(t)}_i g_j\left(\parameters^{(t)}\right) \right]
    \end{align*}
    where we used the fact that, since $\matrixmultipliers^*$ is
    left-stochastic, each of its columns is a
    $\left(\numconstraints+1\right)$-dimensional multinoulli distribution.
    For the second term in the LHS of the second line of
    \eqref{dataset-suboptimality:equilibrium}, we can use the fact that
    $g_1\left(\parameters\right) = 0$:
    \begin{equation*}
      \expectation_{t \sim \indices{T}}\left[ \sum_{i=2}^{\numconstraints+1}
      \multipliers^{(t)}_i g_i\left(\parameters^{(t)}\right) \right] \le
      \sum_{i=2}^{\numconstraints+1} \max_{j \in \indices{\numconstraints+1}}
      \expectation_{t \sim \indices{T}}\left[ \multipliers^{(t)}_i
      g_j\left(\parameters^{(t)}\right) \right]
    \end{equation*}
    Plugging these two results into the second line of
    \eqref{dataset-suboptimality:equilibrium}, the two sums collapse, leaving:
    \begin{equation*}
      \max_{i \in \indices{\numconstraints+1}} \expectation_{t \sim
      \indices{T}}\left[ \multipliers^{(t)}_1 g_i\left(\parameters^{(t)}\right)
      \right] \le \epsilon_{\multipliers}
    \end{equation*}
    Substituting the definitions of $g_i$ and
    $\valgeneralization{\outputParameters}$ (since the $g_i$s are defined on
    $\valdataset$, but we want our result to hold on $\datadistribution$) then
    yields the feasibility claim.
  \end{titled-paragraph}

  \begin{titled-paragraph}{Bound on $\bar{\multipliers}_1$}
    Choosing $\parameters^* = \parameters'$ in
    \eqref{dataset-suboptimality:constraint-lower-bound} (recall that
    $\parameters'$ satisfies all of the proxy constraints with margin
    $\margin$) and substituting the definition of $\bound{\objectiveloss}$:
    \begin{align*}
      \epsilon_{\parameters} + \epsilon_{\multipliers} +
      2\traingeneralization{\Parameters} +
      \valgeneralization{\outputParameters}
      \ge& \expectation_{t \sim \indices{T}, x \sim \datadistribution}\left[
      \multipliers^{(t)}_1 \objectiveloss\left( x; \parameters^{(t)} \right) -
      \multipliers^{(t)}_1 \objectiveloss\left( x; \parameters' \right) \right]
      + \left(1 - \bar{\multipliers}_1\right) \margin \\
      \ge& -\bar{\multipliers}_1 \bound{\objectiveloss} + \left(1 -
      \bar{\multipliers}_1\right) \margin
    \end{align*}
    Solving for $\bar{\multipliers}_1$ yields the claim.
  \end{titled-paragraph}
\end{prf}

Before moving on to the convergence and generalization properties of our actual
algorithms, we need to state some (fairly standard) elementary results:
\begin{definition}
  We say that $\covering \subseteq \R^{\numconstraints + 1}$ is a
  radius-$\coveringradius$ external covering of $\Multipliers \defeq
  \Delta^{\numconstraints+1}$ \wrt the $1$-norm if for every $\multipliers \in
  \Multipliers$ there exists a $\tilde{\multipliers} \in \covering$ for which
  $\norm{\multipliers - \tilde{\multipliers}}_1 \le \coveringradius$.
  Notice that we do not require $\covering$ to be a subset of
  $\Multipliers$---this is why it's an \emph{external} covering.
\end{definition}
\begin{lem}{covering-number}
  Assuming that $\coveringradius \le 1$, there exists a
  radius-$\coveringradius$ external covering of $\Multipliers \defeq \Delta^{\numconstraints+1}$
  \wrt the $1$-norm of size no larger than $\left(5 /
  \coveringradius\right)^{\numconstraints}$.
\end{lem}
\begin{prf}{covering-number}
  Consider the $\numconstraints$-dimensional unit ball $\tilde{B} \defeq
  \left\{ \tilde{\multipliers} \in \R^{\numconstraints} :
  \norm{\tilde{\multipliers}}_1 \le 1 \right\}$ \wrt the $1$-norm (note that we
  could instead consider only the positive orthant, which would improve the
  constant in the overall result). There exists a radius-$\coveringradius / 2$
  covering $\tilde{\covering} \subseteq \R^{\numconstraints}$ of $\tilde{B}$
  with $\abs{\tilde{\covering}} \le \left(1 +
  4/\coveringradius\right)^{\numconstraints} \le
  \left(5/\coveringradius\right)^{\numconstraints}$~\citep{Bartlett:2013}.

  Define $\covering \subseteq \R^{\numconstraints+1}$ as:
  \begin{equation*}
    \covering = \left\{\left[
    \begin{array}{c}
      \tilde{\multipliers} \\
      1 - \norm{\tilde{\multipliers}}_1
    \end{array}
    \right] : \tilde{\multipliers} \in \tilde{\covering} \right\}
  \end{equation*}
  Notice that we do not necessarily have that $\covering \subseteq
  \Delta^{\numconstraints+1}$, \ie this will be an \emph{external} covering.

  From any $\multipliers \in \Delta^{\numconstraints+1}$, we can define
  $\multipliers' \in \R_+^{\numconstraints}$ by dropping the last coordinate of
  $\multipliers$, and we'll have that $\norm{\multipliers'}_1 \le
  \norm{\multipliers}_1 = 1$, so there will exist a $\tilde{\multipliers} \in
  \tilde{\covering}$ such that $\norm{\tilde{\multipliers} - \multipliers'}_1
  \le \coveringradius/2$, which implies that the corresponding element of
  $\covering$ is $\coveringradius$-far from $\multipliers$, showing that
  $\covering$ is a radius-$\coveringradius$ covering of
  $\Delta^{\numconstraints+1}$ \wrt the $1$-norm.
\end{prf}
 \begin{lem}{finite-generalization}
  Let $\dataset$ be an \iid sample from a distribution $\datadistribution$
  supported on $\mathcal{X}$, and $\outputParameters \subseteq \Parameters$ the
  finite set of permitted model parameters, which defines a a finite function
  class ($\outputParameters$ may be a random variable, but must be independent
  of $\dataset$). Suppose that $\loss : \mathcal{X} \times \Parameters
  \rightarrow \left[ a, b \right]$ with $\bound{\loss} \defeq b - a$. Then:
  \begin{equation*}
    \abs{ \frac{1}{\abs{\dataset}} \sum_{x \in \dataset} \loss\left(x;
    \parameters\right) - \expectation_{x \sim \datadistribution}\left[
    \loss\left(x; \parameters\right) \right] }
    < \bound{\loss} \sqrt{ \frac{ \ln \left( 2 \abs{\outputParameters} / \delta
    \right) }{2 \abs{\dataset}} }
  \end{equation*}
  for all $\parameters \in \outputParameters$, with probability at least
  $1-\delta$ over the sampling of $\dataset$.
  \NOTE{the one-sided bound is the same, except that we have
  $\abs{\outputParameters}$ instead of $2 \abs{\outputParameters}$}
\end{lem}
\begin{prf}{finite-generalization}
  Allowing $\outputParameters$ to be a random variable independent of
  $\dataset$, instead of a constant set, doesn't significantly change the
  standard proof~\egcite{Srebro:2016} \TODO{better citation?}.
  By Hoeffding's inequality:
  \begin{equation*}
    \probability\left\{ \abs{ \frac{1}{\abs{\dataset}} \sum_{x \in \dataset}
    \loss\left(x; \parameters\right) - \expectation_{x \sim
    \datadistribution}\left[ \loss\left(x; \parameters\right) \right] } \ge
    \epsilon \right\} \le 2 \exp\left( - \frac{2 \abs{\dataset}
    \epsilon^2}{\bound{\loss}^2} \right)
  \end{equation*}
  the above holding for any $\parameters \in \Parameters$. Since
  $\outputParameters$ is independent of $\dataset$, we can apply the union
  bound over all $\parameters \in \outputParameters$:
  \begin{equation*}
    \probability\left\{ \exists \parameters \in \outputParameters . \abs{
    \frac{1}{\abs{\dataset}} \sum_{x \in \dataset} \loss\left(x;
    \parameters\right) - \expectation_{x \sim \datadistribution}\left[
    \loss\left(x; \parameters\right) \right] } \ge \epsilon \right\} \le 2
    \abs{\parameters} \exp\left( - \frac{2 \abs{\dataset}
    \epsilon^2}{\bound{\loss}^2} \right)
  \end{equation*}
  Rearranging terms yields the claimed result.
\end{prf}

\subsection{\algref{discrete}}\label{app:proofs:discrete}

\begin{lem}{discrete-convergence}
  If we take $\eta_{\multipliers} \defeq \sqrt{\left(\numconstraints+1\right)
  \ln \left(\numconstraints+1\right) / T \bound{\stochasticgrad}^2}$, then the
  result of \algref{discrete} satisfies the conditions of
  \thmref{dataset-suboptimality} with:
  \begin{align*}
    \epsilon_{\parameters} = & \approximation + 2 \coveringradius
    \bound{\tilde{\loss}} \\
    \epsilon_{\multipliers} =& 2 \bound{\stochasticgrad} \sqrt{ \frac{
    \left( \numconstraints + 1 \right) \ln\left( \numconstraints + 1 \right)
    }{T} }
  \end{align*}
  where $\bound{\tilde{\loss}} \ge \abs{ \ell\left(x, \parameters\right) }$ for
  all $\ell \in \left\{ \objectiveloss, \proxyconstraintloss{1}, \dots,
  \proxyconstraintloss{\numconstraints} \right\}$, and $\bound{\stochasticgrad} \ge \max_{t \in
  \indices{T}}\norm{ \stochasticgrad_{\multipliers}^{(t)} }_{\infty}$ is a
  bound on the gradients.
\end{lem}
\begin{prf}{discrete-convergence}
  Since $\covering$ is a radius-$\coveringradius$ external covering of
  $\Multipliers \defeq \Delta^{\numconstraints+1}$ \wrt the $1$-norm, we must
  have that $\norm{ \tilde{\multipliers}^{(t)} - \multipliers^{(t)} }_{1} \le
  \coveringradius$, which implies by \defref{empirical-proxy-lagrangians} that:
  \begin{equation*}
    \abs{ \empiricallagrangian_{\parameters}\left( \parameters,
    \tilde{\multipliers}^{(t)} \right) -
    \empiricallagrangian_{\multipliers}\left( \parameters, \multipliers^{(t)}
    \right) } \le \coveringradius \bound{\tilde{\loss}}
  \end{equation*}
  the above holding for all $\parameters \in \Parameters$, and all $t$. In
  particular, this implies that:
  \begin{align*}
    \empiricallagrangian_{\parameters}\left( \parameters^{(t)},
    \multipliers^{(t)} \right)
    \le &
    \empiricallagrangian_{\parameters}\left( \parameters^{(t)},
    \tilde{\multipliers}^{(t)} \right) + \coveringradius \bound{\tilde{\loss}}
    \\
    \le &
    \inf_{\parameters^* \in
    \Parameters}\empiricallagrangian_{\parameters}\left( \parameters^*,
    \tilde{\multipliers}^{(t)} \right) + \approximation + \coveringradius
    \bound{\tilde{\loss}} \\
    \le &
    \inf_{\parameters^* \in
    \Parameters}\empiricallagrangian_{\parameters}\left( \parameters^*,
    \multipliers^{(t)} \right) + \approximation + 2 \coveringradius
    \bound{\tilde{\loss}}
  \end{align*}
  Therefore:
  \begin{align*}
    \frac{1}{T} \sum_{t=1}^{T} \empiricallagrangian_{\parameters}\left(
    \parameters^{(t)}, \multipliers^{(t)} \right)
    \le & \frac{1}{T} \sum_{t=1}^{T} \inf_{\parameters^* \in \Parameters}
    \empiricallagrangian_{\parameters}\left( \parameters^*, \multipliers^{(t)}
    \right) + \approximation + 2 \coveringradius \bound{\tilde{\loss}} \\
    \le & \inf_{\parameters^* \in \Parameters} \frac{1}{T} \sum_{t=1}^{T}
    \empiricallagrangian_{\parameters}\left( \parameters^*, \multipliers^{(t)}
    \right) + \approximation + 2 \coveringradius \bound{\tilde{\loss}}
  \end{align*}
  so the first condition of \thmref{dataset-suboptimality} is satisfied with
  the claimed $\epsilon_{\parameters}$.

  The second condition, on $\epsilon_{\multipliers}$, follows immediately from
  Lemma 8 of Appendix C.1 of \citet{Cotter:2018}, taking
  $\tilde{\numconstraints} = \numconstraints + 1$.
\end{prf}

\begin{lem}{discrete-generalization}
  If we take $\outputParameters \defeq \left\{ \parameters^{(1)}, \dots,
  \parameters^{(T)} \right\}$ as in \thmref{dataset-suboptimality},
  where $\parameters^{(1)},\dots,\parameters^{(T)}$ are the result of \algref{discrete},
  then with probability $1-\delta$ over the sampling of
  $\valdataset$:
  \begin{equation*}
    \valgeneralization{\outputParameters} < \bound{\loss} \sqrt{ \frac{
    \ln \left( 2 \numconstraints \abs{\covering} / \delta \right) }{2
    \abs{\valdataset}} }
  \end{equation*}
  where $\bound{\loss} \ge \max_{i \in \indices{\numconstraints}} \left( b_i -
  a_i \right)$ assuming that the range of each $\constraintloss{i}$ is the
  interval $\left[ a_i, b_i \right]$.
\end{lem}
\begin{prf}{discrete-generalization}
  Since each $\parameters^{(t)}$ is uniquely associated with a
  $\tilde{\multipliers}^{(t)} \in \covering$ (\defref{oracle}), we will have
  that $\outputParameters \subseteq \coveringParameters$, where:
  \begin{equation*}
    \coveringParameters \defeq \left\{ \oracle\left(
    \empiricallagrangian_{\parameters}\left(\cdot,\tilde{\multipliers}\right)
    \right) : \tilde{\multipliers} \in \covering \right\}
  \end{equation*}
  Because the oracle call defining $\coveringParameters$ depends only on
  $\empiricallagrangian_{\parameters}$, which itself depends only on
  $\traindataset$, we can apply \lemref{finite-generalization} to
  $\coveringParameters$, yielding that, for each $i \in
  \indices{\numconstraints}$, the following holds with probability $\delta /
  \numconstraints$ for all $\parameters \in \coveringParameters$:
  \begin{equation*}
    \abs{ \frac{1}{\abs{\valdataset}} \sum_{x \in \valdataset}
    \constraintloss{i}\left(x; \parameters\right) - \expectation_{x \sim
    \datadistribution}\left[ \constraintloss{i}\left(x; \parameters\right)
    \right] }
    < \bound{\loss} \sqrt{ \frac{ \ln \left( 2 \numconstraints
    \abs{\coveringParameters} / \delta \right) }{2 \abs{\valdataset}} }
  \end{equation*}
  The claimed result on $\valgeneralization{\outputParameters}$ then follows
  from the union bound and the facts that $\outputParameters \subseteq
  \coveringParameters$ and $\abs{\coveringParameters} = \abs{\covering}$.
\end{prf}

\subsection{\algref{continuous}}\label{app:proofs:continuous}

\begin{lem}{continuous-convergence}
  Suppose that $\Parameters$ is compact and convex, and that $\loss\left(x;
  \parameters\right)$ is $\strongconvexity$-strongly convex in $\parameters$
  for all $\loss \in \left\{ \objectiveloss, \proxyconstraintloss{1}, \dots,
  \proxyconstraintloss{\numconstraints} \right\}$.
  If we take $\eta_{\multipliers} \defeq \sqrt{\left(\numconstraints+1\right)
  \ln \left(\numconstraints+1\right) / T_{\multipliers}
  \bound{\stochasticgrad}^2}$, then the result of \algref{continuous}
  satisfies the conditions of \thmref{dataset-suboptimality} with:
  \begin{align*}
    \epsilon_{\parameters} = & \frac{\bound{\stochasticsubgrad}^2 \left(1 + \ln
    T_{\parameters}\right)}{2 \strongconvexity T_{\parameters}} \\
    \epsilon_{\multipliers} =& 2 \bound{\stochasticgrad} \sqrt{ \frac{ \left(
    \numconstraints + 1 \right) \ln\left( \numconstraints + 1 \right)
    }{T_{\multipliers}} }
  \end{align*}
  where $\bound{\stochasticsubgrad} \ge \max_{s,t \in \indices{T_{\parameters}}
  \times \indices{T_{\multipliers}}}\norm{
  \stochasticsubgrad_{\parameters}^{(t,s)} }_{2}$ is a bound on the
  subgradients, and $\bound{\stochasticgrad} \ge \max_{t \in
  \indices{T_{\multipliers}}}\norm{ \stochasticgrad_{\multipliers}^{(t)}
  }_{\infty}$ is a bound on the gradients.
\end{lem}
\begin{prf}{continuous-convergence}
  By Lemma 1 of \citet{ShalevShwartz:2010}, the fact that
  $\empiricallagrangian_{\parameters}\left(\parameters, \multipliers\right)$ is
  $\strongconvexity$-strongly convex in $\parameters$ (because $\loss\left(x;
  \parameters\right)$ is $\strongconvexity$-strongly convex in $\parameters$
  for $\loss \in \left\{ \objectiveloss, \proxyconstraintloss{1}, \dots,
  \proxyconstraintloss{\numconstraints} \right\}$, and $\multipliers \in
  \Multipliers \defeq \Delta^{\numconstraints+1}$), and Jensen's inequality:
  \begin{equation}
    \label{eq:lem:continuous-convergence:parameters-suboptimality}
    \empiricallagrangian_{\parameters}\left( \parameters^{(t)},
    \multipliers^{(t)} \right)
    \le
    \frac{1}{T_{\parameters}} \sum_{s=1}^{T_{\parameters}}
    \empiricallagrangian_{\parameters}\left( \tilde{\parameters}^{(t,s)},
    \multipliers^{(t)} \right)
    \le
    \min_{\parameters^* \in \Parameters}
    \empiricallagrangian_{\parameters}\left( \parameters^*, \multipliers^{(t)}
    \right) + \frac{\bound{\stochasticsubgrad}^2 \left(1 + \ln
    T_{\parameters}\right)}{2 \strongconvexity T_{\parameters}}
  \end{equation}
  the above holding for all $t$. Therefore:
  \begin{equation*}
    \frac{1}{T} \sum_{t=1}^{T} \empiricallagrangian_{\parameters}\left(
    \parameters^{(t)}, \multipliers^{(t)} \right)
    \le
    \min_{\parameters^* \in \Parameters} \frac{1}{T} \sum_{t=1}^{T}
    \empiricallagrangian_{\parameters}\left( \parameters^*, \multipliers^{(t)}
    \right) + \frac{\bound{\stochasticsubgrad}^2 \left(1 + \ln
    T_{\parameters}\right)}{2 \strongconvexity T_{\parameters}}
  \end{equation*}
  so the first condition of \thmref{dataset-suboptimality} is satisfied with
  the claimed $\epsilon_{\parameters}$.

  As in the proof of \lemref{discrete-convergence}, the second condition, on
  $\epsilon_{\multipliers}$, follows immediately from Lemma 8 of Appendix C.1
  of \citet{Cotter:2018}, taking $\tilde{\numconstraints} = \numconstraints +
  1$.
\end{prf}

\begin{lem}{continuous-generalization}
  In addition to the conditions of \lemref{continuous-convergence}, suppose
  that $\loss\left(x; \parameters\right)$ is $\lipschitz$-Lipschitz continuous
  in $\parameters$ for all $\loss \in \left\{ \constraintloss{1}, \dots,
  \constraintloss{\numconstraints} \right\}$.
  If we take $\outputParameters \defeq \left\{ \parameters^{(1)}, \dots,
  \parameters^{(T)} \right\}$ as in \thmref{dataset-suboptimality}, where
  $\parameters^{(1)},\dots,\parameters^{(T)}$ are the result of
  \algref{continuous}, then with probability $1-\delta$ over the sampling of
  $\valdataset$:
  \begin{equation*}
    \valgeneralization{\outputParameters} <
    2 \lipschitz \sqrt{ \frac{4 \coveringradius
    \bound{\tilde{\loss}}}{\strongconvexity} } +
    \frac{2 \lipschitz \bound{\stochasticsubgrad}}{\strongconvexity} \sqrt{ \frac{1 + \ln
    T_{\parameters}}{T_{\parameters}} }
    + \bound{\loss} \sqrt{ \frac{ \ln \left( 2 \numconstraints \abs{\covering}
    / \delta \right) }{2 \abs{\valdataset}} }
  \end{equation*}
  where $\bound{\tilde{\loss}} \ge \abs{ \ell\left(x, \parameters\right) }$ for
  all $\ell \in \left\{ \objectiveloss, \proxyconstraintloss{1}, \dots,
  \proxyconstraintloss{\numconstraints} \right\}$, $\bound{\loss} \ge \max_{i \in
  \indices{\numconstraints}} \left( b_i - a_i \right)$ assuming that the range
  of each $\constraintloss{i}$ is the
  interval $\left[ a_i, b_i \right]$, and $\covering$ is a
  radius-$\coveringradius$ covering of $\Multipliers \defeq
  \Delta^{\numconstraints+1}$ \wrt the $1$-norm.
\end{lem}
\begin{prf}{continuous-generalization}
  Define $\tilde{\multipliers}^{(t)} \defeq \argmin_{\tilde{\multipliers} \in
  \covering} \norm{\multipliers^{(t)} - \tilde{\multipliers}}$ for all $t$.
  Since $\covering$ is a radius-$\coveringradius$ covering of $\Multipliers
  \defeq \Delta^{\numconstraints+1}$ \wrt the $1$-norm, we must have that
  $\norm{ \tilde{\multipliers}^{(t)} - \multipliers^{(t)} }_{1} \le
  \coveringradius$, which implies by \defref{empirical-proxy-lagrangians} that:
  \begin{equation*}
    \abs{ \empiricallagrangian_{\parameters}\left( \parameters,
    \tilde{\multipliers}^{(t)} \right) -
    \empiricallagrangian_{\multipliers}\left( \parameters, \multipliers^{(t)}
    \right) } \le \coveringradius \bound{\tilde{\loss}}
  \end{equation*}
  the above holding for all $\parameters \in \Parameters$, and all $t$.
  Take $\parameters^{(t*)} \defeq \argmin_{\parameters^* \in \Parameters}
  \empiricallagrangian_{\parameters}\left(\parameters^*,
  \multipliers^{(t)}\right)$ and $\tilde{\parameters}^{(t*)} \defeq
  \argmin_{\tilde{\parameters}^* \in \Parameters}
  \empiricallagrangian_{\parameters}\left(\tilde{\parameters}^*,
  \tilde{\multipliers}^{(t)}\right)$. Then, by the above result and the
  triangle inequality:
  \begin{align*}
    \abs{ \empiricallagrangian_{\parameters}\left(\parameters^{(t*)},
    \multipliers^{(t)}\right) -
    \empiricallagrangian_{\parameters}\left(\tilde{\parameters}^{(t*)},
    \tilde{\multipliers}^{(t)}\right) } \le & \coveringradius
    \bound{\tilde{\loss}} \\
    \abs{ \empiricallagrangian_{\parameters}\left(\parameters^{(t*)},
    \tilde{\multipliers}^{(t)}\right) -
    \empiricallagrangian_{\parameters}\left(\tilde{\parameters}^{(t*)},
    \tilde{\multipliers}^{(t)}\right) } \le & 2 \coveringradius
    \bound{\tilde{\loss}}
  \end{align*}
  so by the fact that $\empiricallagrangian_{\parameters}\left(\parameters,
  \multipliers\right)$ is $\strongconvexity$-strongly convex in $\parameters$
  for all $\multipliers$:
  \begin{equation*}
    \norm{ \parameters^{(t*)} - \tilde{\parameters}^{(t*)} }_2 \le \sqrt{
      \frac{4 \coveringradius \bound{\tilde{\loss}}}{\strongconvexity} }
  \end{equation*}
  Again by strong convexity, but applied to
  \eqref{lem:continuous-convergence:parameters-suboptimality} in the proof of
  \lemref{continuous-convergence}:
  \begin{equation*}
    \norm{ \parameters^{(t)} - \parameters^{(t*)} }_2 \le \sqrt{
      \frac{\bound{\stochasticsubgrad}^2 \left(1 + \ln
      T_{\parameters}\right)}{\strongconvexity^2 T_{\parameters}} }
  \end{equation*}
  so by the triangle inequality:
  \begin{equation*}
    \norm{ \parameters^{(t)} - \tilde{\parameters}^{(t*)} }_2 \le \sqrt{
      \frac{4 \coveringradius \bound{\tilde{\loss}}}{\strongconvexity} } +
      \sqrt{ \frac{\bound{\stochasticsubgrad}^2 \left(1 + \ln
      T_{\parameters}\right)}{\strongconvexity^2 T_{\parameters}} }
  \end{equation*}
  and by $\lipschitz$-Lipschitz continuity:
  \begin{equation}
    \label{eq:lem:continuous-generalization:lipschitz}
    \abs{ \constraintloss{i}\left(x; \parameters^{(t)}\right) -
    \constraintloss{i}\left(x; \tilde{\parameters}^{(t*)}\right) } \le
    \lipschitz\left( \sqrt{ \frac{4 \coveringradius
    \bound{\tilde{\loss}}}{\strongconvexity} } + \sqrt{
      \frac{\bound{\stochasticsubgrad}^2 \left(1 + \ln
      T_{\parameters}\right)}{\strongconvexity^2 T_{\parameters}} } \right)
  \end{equation}
  the above holding for all $t$, and all $i\in\indices{\numconstraints}$.

  Define $\tilde{\Parameters} \defeq \left\{ \tilde{\parameters}^{(1*)}, \dots,
  \tilde{\parameters}^{(T*)} \right\}$. Observe that since
  $\tilde{\parameters}^{(t*)}$ is uniquely associated with a
  $\tilde{\multipliers}^{(t)} \in \covering$, we will have that
  $\tilde{\Parameters} \subseteq \coveringParameters$, where:
  \begin{equation*}
    \coveringParameters \defeq \left\{ \argmin_{\tilde{\parameters}^* \in
    \Parameters} \empiricallagrangian_{\parameters}\left(\tilde{\parameters}^*,
    \tilde{\multipliers}\right) : \tilde{\multipliers} \in \covering \right\}
  \end{equation*}
  Because the $\argmin$s defining $\coveringParameters$ depend only on
  $\empiricallagrangian_{\parameters}$, which itself depends only on
  $\traindataset$, we can apply \lemref{finite-generalization} to
  $\coveringParameters$, yielding that, for each $i \in
  \indices{\numconstraints}$, the following holds with probability $\delta /
  \numconstraints$ for any $\parameters \in \coveringParameters$:
  \begin{equation*}
    \abs{ \frac{1}{\abs{\valdataset}} \sum_{x \in \valdataset}
    \constraintloss{i}\left(x; \parameters\right) - \expectation_{x \sim
    \datadistribution}\left[ \constraintloss{i}\left(x; \parameters\right)
    \right] }
    < \bound{\loss} \sqrt{ \frac{ \ln \left( 2 \numconstraints
    \abs{\coveringParameters} / \delta \right) }{2 \abs{\valdataset}} }
  \end{equation*}
  By the union bound, we could instead take the above to hold uniformly for all
  $i \in \indices{\numconstraints}$ with probability $1-\delta$. Substituting
  \eqref{lem:continuous-generalization:lipschitz} and using the facts that
  $\tilde{\Parameters} \subseteq \coveringParameters$ and
  $\abs{\coveringParameters} = \abs{\covering}$ yields the claimed result.
\end{prf}

\section{One-dataset Lagrangian Baseline Approach}\label{app:baseline}

In this appendix, we'll analyze the most natural theoretical baseline for our
proposed approach, namely using a single training dataset, and optimizing the
empirical Lagrangian:
\begin{equation}
  \label{eq:empirical-lagrangian}
  \empiricallagrangian\left(\parameters, \multipliers\right) \defeq
  \frac{1}{\abs{\traindataset}} \sum_{x \in \traindataset} \left(
  \objectiveloss\left(x;\parameters\right) + \sum_{i=1}^{\numconstraints}
  \multipliers_i \constraintloss{i}\left(x;\parameters\right) \right)
\end{equation}
This is essentially an extension of the approach proposed by
\citet{Agarwal:2018}---who proposed using the Lagrangian formulation in the
particular case of fair classification---to the slightly more general setting
of inequality constrained optimization.

Theorem 1 of \citet{Cotter:2018} characterizes the optimality and feasibility
properties of Nash equilibria of the in-expectation Lagrangian
(\eqref{lagrangian}). The analogous result for the empirical Lagrangian of
\eqref{empirical-lagrangian} is given in the following theorem:
\begin{thm}{lagrangian-dataset-suboptimality}
  Define $\Multipliers = \left\{ \multipliers \in \R_+^{\numconstraints} :
  \norm{\multipliers}_1 \le \Radius \right\}$, let $\proxyconstraintloss{i}
  \defeq \constraintloss{i}$ for all $i\in\indices{\numconstraints}$, and
  consider the empirical Lagrangian of \eqref{empirical-lagrangian}.
  Let $\parameters^{(1)},\dots,\parameters^{(T)} \in \Parameters$ and
  $\multipliers^{(1)},\dots,\multipliers^{(T)} \in \Multipliers$ be sequences
  satisfying:
  \begin{equation}
    \max_{\multipliers^* \in \Multipliers} \frac{1}{T} \sum_{t=1}^T
    \empiricallagrangian\left( \parameters^{(t)}, \multipliers^* \right) -
    \inf_{\parameters^* \in \Parameters} \frac{1}{T} \sum_{t=1}^T
    \empiricallagrangian\left( \parameters^*, \multipliers^{(t)} \right) \le
    \epsilon
  \end{equation}
  Define $\bar{\parameters}$ as a random variable for which $\bar{\parameters}
  = \parameters^{(t)}$ with probability $1/T$, and let $\bar{\multipliers}
  \defeq \left(\sum_{t=1}^T \multipliers^{(t)}\right) / T$.
  Then $\bar{\parameters}$ is nearly-optimal in expectation:
  \begin{equation*}
    \expectation_{\bar{\parameters}, x \sim \datadistribution}\left[
    \objectiveloss\left(x; \bar{\parameters}\right) \right] \le
    \inf_{\parameters^* \in \Parameters : \forall i .  \expectation_{x \sim
    \datadistribution}\left[ \constraintloss{i}\left(x; \parameters^*\right)
    \right] \le 0} \expectation_{x \sim \datadistribution}\left[
    \objectiveloss\left(x; \parameters^*\right) \right] + \epsilon
    + 2 \traingeneralization{\Parameters}
  \end{equation*}
  and nearly-feasible:
  \begin{equation*}
    \max_{i \in \indices{\numconstraints}} \expectation_{\bar{\parameters}, x
    \sim \datadistribution}\left[ \constraintloss{i}\left(x;
    \bar{\parameters}\right) \right] \le \frac{\epsilon}{\Radius -
    \norm{\bar{\multipliers}}_1} + \traingeneralization{\Parameters}
  \end{equation*}
  Additionally, if there exists a $\parameters' \in \Parameters$ that satisfies
  all of the constraints with margin $\margin$ (\ie $\expectation_{x \sim
  \datadistribution}\left[ \constraintloss{i}\left(x; \parameters'\right)
  \right] \le -\margin$ for all $i \in \indices{\numconstraints}$), then:
  \begin{equation*}
    \norm{ \bar{\multipliers} }_1 \le \frac{\epsilon +
    \bound{\objectiveloss}}{\margin - \traingeneralization{\Parameters}}
  \end{equation*}
  assuming that $\margin > \traingeneralization{\Parameters}$,
  where $\bound{\objectiveloss} \ge \sup_{\parameters \in \Parameters}
  \expectation_{x \sim \datadistribution}\left[ \objectiveloss\left(x;
  \parameters\right) \right] - \inf_{\parameters \in \Parameters}
  \expectation_{x \sim \datadistribution}\left[ \objectiveloss\left(x;
  \parameters\right) \right]$ is a bound on the range of the objective loss.
\end{thm}
\begin{prf}{lagrangian-dataset-suboptimality}
  %
  The empirical Lagrangian is nothing but the in-expectation Lagrangian over
  the finite training sample $\traindataset$, so by Theorem 1 of
  \citet{Cotter:2018}, $\bar{\parameters}$ is nearly-optimal in expectation:
  \begin{equation*}
    \expectation_{\bar{\parameters}, x \sim \traindataset}\left[
    \objectiveloss\left(x; \bar{\parameters}\right) \right] \le
    \inf_{\parameters^* \in \Parameters : \forall i .  \expectation_{x \sim
    \traindataset}\left[ \constraintloss{i}\left(x; \parameters^*\right)
    \right] \le 0} \expectation_{x \sim \traindataset}\left[
    \objectiveloss\left(x; \parameters^*\right) \right] + \epsilon
  \end{equation*}
  and nearly-feasible:
  \begin{equation*}
    \max_{i \in \indices{\numconstraints}} \expectation_{\bar{\parameters}, x
    \sim \traindataset}\left[ \constraintloss{i}\left(x;
    \bar{\parameters}\right) \right] \le \frac{\epsilon}{\Radius -
    \norm{\bar{\multipliers}}_1}
  \end{equation*}
  Since $\parameters'$ satisfies the constraints with margin $\margin$ in
  expectation, it will satisfy them with margin $\margin -
  \traingeneralization{\Parameters}$ on the training dataset, so the same
  theorem gives the claimed upper-bound $\norm{\bar{\multipliers}}_1 \le
  \left(\epsilon + \bound{\objectiveloss}\right) / (\margin -
  \traingeneralization{\Parameters})$ when $\margin >
  \traingeneralization{\Parameters}$.

  Notice that the above expectations are taken over the finite training sample
  $\traindataset$, rather than the data distribution $\datadistribution$. To
  fix this, we need only define $\proxyconstraintloss{i} = \constraintloss{i}$,
  and appeal to the definition of $\traingeneralization{\Parameters}$
  (\defref{generalization}), yielding the claimed results.
\end{prf}

Here, $\Radius$ is the maximum allowed $1$-norm of the vector of Lagrange
multipliers (such a bound is necessary for \citet{Cotter:2018}'s proof to work
out).
Notice that we have assumed that $\proxyconstraintloss{i} \defeq
\constraintloss{i}$ for all $i$. This is purely for notational reasons (the
Lagrangian does not involve proxy constraints \emph{at all})---it allows us to
re-use the definition of $\traingeneralization{\Parameters}$ in the above
Theorem, and causes $\margin$ to have the same definition here, as it did in
\appref{proofs}.

\end{document}